\documentclass{article}
\usepackage{ijcai20}

\usepackage{times}
\usepackage{enumitem}
\usepackage{soul}
\usepackage{url}
\usepackage[hidelinks]{hyperref}
\usepackage[utf8]{inputenc}
\usepackage[small]{caption}
\usepackage{graphicx}
\usepackage{amsmath}
\usepackage{amsthm}
\usepackage{booktabs}
\usepackage{algorithm}
\usepackage{algorithmic}
\usepackage[dvipsnames]{xcolor}
\usepackage{amssymb}
\usepackage{booktabs}
\usepackage{algorithm}
\usepackage{pdfpages}
\usepackage{algorithmic}
\usepackage[toc,page]{appendix}
\usepackage{tabulary}
\newtheorem{definition}{Definition}
\newtheorem{theorem}{Theorem}
\newtheorem{lemma}{Lemma}
\usepackage{subcaption}
\usepackage[hidelinks]{hyperref}
\title{FNNC: Achieving Fairness through Neural Networks}
\author{
Padala Manisha
\and
Sujit Gujar
\affiliations
International Institute of Information Technology, Hyderabad
\emails
manisha.padala@research.iiit.ac.in,
sujit.gujar@iiit.ac.in
}
\begin{document}

\maketitle
% \includepdf[pages=-]{CoverLetter}
% \includepdf[pages=-]{1582}
% \includepdf[pages=-]{IJCAI_submission}

\begin{abstract}
In classification models fairness can be ensured by solving a constrained optimization problem. We focus on fairness constraints like Disparate Impact, Demographic Parity, and Equalized Odds, which are non-decomposable and non-convex. Researchers define convex surrogates of the constraints and then apply convex optimization frameworks to obtain fair classifiers. Surrogates serve only as an upper bound to the actual constraints, and convexifying fairness constraints might be challenging.

We propose a neural network-based framework, \emph{FNNC}, to achieve fairness while maintaining high accuracy in classification. The above fairness constraints are included in the loss using Lagrangian multipliers. We prove bounds on generalization errors for the constrained losses which asymptotically go to zero. The network is optimized using two-step mini-batch stochastic gradient descent. Our experiments show that FNNC performs as good as the state of the art, if not better.  The experimental evidence supplements our theoretical guarantees. In summary, we have an automated solution to achieve fairness in classification, which is easily extendable to many fairness constraints.

\end{abstract}

\maketitle
\section{Introduction}

In recent years machine learning models have been popularized as prediction models to supplement the process of decision making. Such models are used for criminal risk assessment, credit approvals, online advertisements.  These machine learning models unknowingly introduce a societal bias through their predictions \cite{barocas16,berk,chouldechova17}. E.g., ProPublica conducted its study of the risk assessment tool, which was widely used by the judiciary system in the USA. ProPublica observed that the risk values for recidivism estimated for African-American defendants were on an average higher than for Caucasian defendants. Since then, researchers started looking at fairness in machine learning, especially quantifying the notion of fairness and achieving it.  

Broadly fairness measures are divided into two categories. \emph{Individual fairness} \cite{dwork12},  requires similar decision outcomes for two individuals belonging to two different groups concerning the sensitive feature and yet sharing similar non-sensitive features. The other notion is of \emph{group fairness} \cite{zemel13}, which requires different sensitive groups to receive beneficial outcomes in similar proportions. We are concerned with group fairness and specifically: \emph{Demographic Parity} (DP) \cite{dwork12}, \emph{Disparate Impact} (DI) \cite{feldman15} and \emph{Equalized odds} (EO) \cite{dawid82,hardt16}. DP ensures that the fraction of the positive outcome is the same for all the groups. DI ensures the ratio of the fractions is above a threshold. However, both the constraints fail when the base rate itself differs, hence EO is the more useful notion of fairness, which ensures even distribution of false-positive rates and false-negative rates among the groups. All these definitions make sense only when the classifier is well calibrated. That is, if a classifier predicts an instance belongs to a class with a probability of $0.8$, then there should be $80\%$ of samples belonging to that class. \cite{raghavan17,chouldechova17} show that it is impossible to achieve EO with calibration unless we have perfect classifiers. Hence, the major challenge is to devise an algorithm that guarantees the best predictive accuracy while satisfying the fairness constraints to a certain degree. 

Towards designing such algorithms, one approach is pre-processing the data. The methods under this approach treat the classifier as a black box and focus on learning fair representations. The fair representations learned may not result in optimal accuracy. The other approach models achieving fairness as constrained optimization, \cite{zafar17,kamishima11,fairness_aware_18}. In  \cite{fairness_aware_18}, the authors have provided a generalized convex optimization framework with theoretical guarantees. The fairness constraints are upper-bounded by convex surrogate functions and then directly incorporated into classification models. 

There are several limitations in the existing approaches which ensure fairness in classification models.
Surrogate constraints may not be a reasonable estimate of the original fairness constraint.  Besides, coming up with good surrogate losses for the different definitions of fairness is challenging. In this paper, we study how to achieve fairness in classification. In doing so, we do not aim to propose a new fairness measure or new optimization technique. As opposed to the above approaches, we propose to use neural networks for implementing non-convex complex measures like DP, DI, or EO. The network serves as a simple classification model that achieves fairness. One need not define surrogates or do rigorous analysis to design the model. Mainly, it is adaptable to any definition of fairness. 

Typically, one cannot evaluate fairness measures per sample as these measures make sense only when calculated across a batch, which contains data points from all the sensitive groups. Given that
at every iteration, the network processes mini-batch of data,
we can approximate the fairness measure given an appropriate
batch size.  Hence, we use mini-batch stochastic gradient descent (SGD) for optimizing the network. We empirically find that it is possible to train a network using the Lagrangian Multiplier method, which ensures these constraints and achieves accuracy at par with the other complex frameworks. Likewise, it is also possible to incorporate other complex measures like F1-score,
H-mean loss, and Q-mean loss,  -- not related to fairness.
We have included an experiment on training a network to
minimize Q-mean loss with DP as a constraint.

\noindent \emph{Our Contribution:} i) We propose to design a fair neural network classifier (FNNC) to achieve fairness in classification.  ii) We provide generalization bounds for the different losses and fairness constraints DP and EO (Theorem \ref{thm:gb_2}) in FNNC. iii) We show that, in some instances, it may be difficult to approximate DI constraint by another surrogate DI constraint (Theorem \ref{thm:gb_3}). iv) We empirically show that FNNC can achieve the state of the art performance, if not better.

%%%%%%%%%%%%%%%%%%%%%%%%%%%%%%%%%%%%%%%%%%%%%%%%%%%%%%%%%%%%%%%%%%%%%%%%%%%%%%%%
\section{Related Work} 
\label{sec:rw}
In \cite{zemel13}, the notion of fairness is discussed. DP, EO, and DI are few of its types. It is a major challenge to enforce these in any general machine learning framework. Widely there are three primary approaches to deal with the challenge: 

i) The first body of work focuses on pre-processing i.e., coming up with fair representations as opposed to fair classification e.g., \cite{feldman15,dwork12,kamiran09,kamiran10}. Neural networks have been extensively used in such pursuit. E.g., \cite{louizos15} gives a method for learning fair representations with a variational auto-encoder by using maximum mean discrepancies between the two sensitive groups.\cite{edwards16,madras18,beutel17}  explore the notion of adversarially learning a classifier that achieves DP, EO or DI.

ii) The second approach focuses on analytically designing convex surrogates for the fairness definitions \cite{calders10,kamishima11,bechavod17} introduce penalty functions to penalize unfairness. \cite{zafar17,fairness_aware_18} gives a generalized convex framework that incorporates all possible surrogates and gives appropriate bounds. \cite{zhang18} uses neural network-based adversarial learning, which attempts to predict the sensitive attribute based on the classifier output, to learn an equal opportunity classifier. 

iii) The third is the reductionist approach, in which the task of fair classification is reduced to a sequence of cost-sensitive classification \cite{narasimhan18}, and \cite{agarwal18} which can then be solved by a standard classifier. 
\cite{agarwal18} allows for fairness definitions that can be characterized as linear inequalities under conditional moments like DP and EO (DI does not qualify for the same). FNNC does not have such restrictions and hence performs reasonably for DI as well. We are easily able to include complex and non-decomposable loss functions like Q-mean loss, whereas \cite{agarwal18} aims to improve only the accuracy of the model.

%%%%%%%%%%%%%%%%%%%%%%%%%%%%%%%%%%%%%%%%%%%%%%%%%%%%%%%%%%%%%%%%%%%%%%%%%%%%%%%%%%

\section{Preliminaries and Background}
\label{sec:prem}
In this section, we introduce the notation used and state the definitions of the fairness measures and the performance measures that we have analyzed.

We consider a binary classification problem with no assumption on the instance space. $X$ is our  ($d$-dimensional) instance space s.t. $X \in \mathbb{R}^d$ and output space $Y \in \{0,1\}$. We also have a protected attribute $\mathcal{A}$ associated with each individual instance, which for example could be age, sex or caste information. For each $a \in \mathcal{A}$, $a$ could be a particular category of the sensitive attribute like male or female.

\begin{definition}[Demographic Parity (DP)] \label{def:dp} A classifier $h$ satisfies demographic parity under a distribution over $(X, \mathcal{A}, Y)$ if its predictions $h(X)$ is independent of the protected attribute $A$. That is, $\forall a \in \mathcal{A} \ and \ p \in \{0,1\}$
$$ \mathbf{P}[h(X)=p | \mathcal{A} = a] = \mathbf{P}[h(X)=p]$$
Given that $p \in \{0,1\}$, we can say $\forall \ a$ $$\mathbb{E}[h(X) | \mathcal{A} = a] = \mathbb{E}[h(X)] $$ 
\end{definition}

\begin{definition}[Equalized Odds (EO)] A classifier $h$ satisfies equalized odds under a distribution over $(X, \mathcal{A}, Y)$ if its predictions $h(X)$ are independent of the protected attribute $\mathcal{A}$ given the label $Y$. That is, $\forall a \in \mathcal{A}, \ p \in \{0,1\} \ and \ y \in Y$
$$ \mathbf{P}[h(X)=p | \mathcal{A} = a, Y = y] = \mathbf{P}[h(X)=p | Y=y]$$
Given that $p \in \{0,1\}$, we can say $\forall \ a, y$ $$\mathbb{E}[h(X) | \mathcal{A} = a, Y=y] = \mathbb{E}[h(X) | Y=y] $$ 
\end{definition}

\begin{definition}[Disparate Impact (DI)] The outcomes of a classifier $h$ disproportionately hurt people with certain sensitive attributes. The following is the definition for completely removing DI,
$$min \bigg( \frac{\mathbf{P}(h(x)>0 | a=1)}{\mathbf{P}(h(x)>0 | a=0)}, \frac{\mathbf{P}(h(x)>0 | a=0)}{\mathbf{P}(h(x)>0 | a=1)} \bigg) = 1$$
\end{definition}

\cite{raghavan17} strongly claim that the above mentioned measures are rendered useless, if the classifier is not calibrated, in which case the probability estimate $p$ of the classifier could carry different meanings for the different groups.
\begin{definition}[Calibration]
A classifier $h$ is perfectly calibrated if $\forall \ p \in [0,1], \ \mathbf{P}(y = 1 | h(x)=p) = p.$
\end{definition}
Given the definition the authors prove the following impossibility of calibration with equalized odds.
\begin{theorem} [Impossibility Result \cite{raghavan17}]
Let $h_1$, $h_2$ be two classifiers for groups $a_1$ and $a_2$ $\in \mathcal{A}$ 
with $\mathbf{P}(y=1|a_1=1) \neq \mathbf{P}(y=1|a_2=1)$. Then $h_1$ and $h_2$ satisfy the equalized odds and calibration constraints if and only if $h_1$ and $h_2$ are perfect predictors.
\end{theorem}
Given the above result, we cannot guarantee to ensure the fairness constraints perfectly, hence we relax the conditions while setting up our optimization problem as follows,

\subsubsection{Problem Framework}
\label{subsec:problem} We have used the cross-entropy loss or the Q-mean loss as our performance measures, defined in the next section. We denote this loss by $l(h_{\theta}(X), Y)$ parameterized by $\theta$, the weights of the network. Our aim is to minimize the loss under the additional constraints of fairness. Below we state the $\epsilon$-relaxed fairness constraints that we implement in our model. $\forall \ a, y$, 

 \noindent DP: \begin{equation}
\label{eq:dp}
|\mathbb{E}[h(X=x) | \mathcal{A} = a] - \mathbb{E}[h(X=x)]| \leq \epsilon
\end{equation} 
 EO: \begin{equation} \label{eq:eo}
|\mathbb{E}[h(X=x) | \mathcal{A} = a, Y=y] - \mathbb{E}[h(X=x)|Y=y]| \leq \epsilon
\end{equation} 
 DI:  It is not possible to completely remove DI but one has to ensure least possible DI specified by the $p\%-rule$, 
\begin{equation} \label{eq:di}
  min \bigg( \frac{\mathbf{P}(h(x)>0 | a=1)}{\mathbf{P}(h(x)>0 | a=0)}, \frac{\mathbf{P}(h(x)>0 | a=0)}{\mathbf{P}(h(x)>0 | a=1)} \bigg) \geq \frac{p}{100}
\end{equation}

We have the following generic optimization framework. Both the loss and the constraints can be replaced according to the need, 
\begin{equation}
\label{eq:opt}
\boxed{
\begin{aligned}
\underset{\theta}{min}\ l_{\theta} \\
s.t. \  Eq \ \ref{eq:dp} \ or \ & \ref{eq:eo} \ or  \ \ref{eq:di} 
\end{aligned}
}
\end{equation}

%%%%%%%%%%%%%%%%%%%%%%%%%%%%%%%%%%%%%%%%%%%%%%%%%%%%%%%%%%%%%%%%%%%%%%%%%%%%%%%%%

\section{FNNC and Loss Functions}
\label{sec:nn}
In this section, we discuss how we use the neural network for solving the optimization problem framework in Eq. \ref{eq:opt}. 
\subsection{Network Architecture}
Our network is a two-layered feed-forward neural network. We only consider binary classification in all our experiments, although this method and the corresponding definitions are easily extendable to multiple classes.
Let $h_{\theta}(.)$ be the function parameterized by $\theta$ that the neural network learns. In the last layer of this network we have a softmax function which gives the prediction probability $p_{i}$, where $p_i$ is the predicted probability that the $i^{th}$ data sample belongs to one class and $1 - p_i$ is the probability for that it belongs to the other. Hence $p := h_{\theta}(.)$. We use the output probabilities to define the loss and the fairness measure.  
\subsection{Loss function and Optimizer}
Given the constrained optimization defined by Eq. \ref{eq:opt}, we use the Lagrangian Multiplier method to incorporate the constraints within a single overall loss. Since the constraints are non-convex, we can only guarantee that the optimizer converges to a local minima. Nevertheless, our experiments show that the model has at par or better performance compared to the existing approaches. We now describe the different loss functions that we have used in the experiments. 
\subsubsection{Fairness constraints with cross entropy loss:}
The fairness constraint DP as in the Def. \ref{def:dp} is given by $\forall a \in \mathcal{A}$,
\begin{equation*}
  \begin{split}
    \mathbb{E}[h(X=x)| \mathcal{A}=a] &= \mathbb{E}[h(X=x)]  \\ 
    \mathbb{E}[h(X=x)| \mathcal{A}= 1-a] &= \mathbb{E}[h(X=x)]  \\ 
    \therefore\ \mathbb{E}[h(X=x)| \mathcal{A}=a] &= \mathbb{E}[h(X=x)| \mathcal{A}= 1-a]  \\ 
\end{split}
\end{equation*}

Hence the constraint for a fixed batch size $S$ of samples given by $z_S=(h(x_S), a_S, y_S)$ and  $p_i = h(x_i) \in [0,1]$, can be defined as follows,
$$const^{DP}(z_S) = \bigg|\frac{\sum_{i=1}^S p_i a_i}{\sum_{i=1}^S a_i} - \frac{\sum_{i=1}^S p_i (1 - a_i)}{\sum_{i=1}^S 1-a_i}\bigg|$$

For the next constraint EO, we first define the difference in false-positive rate between the two sensitive attributes,
$$ fpr(z_S) =  \bigg| \frac{\sum_{i=1}^S p_i (1 - y_i) a_i}{\sum_{i=1}^S a_i} - \frac{\sum_{i=1}^S p_i(1 -  y_i)(1 - a_i)}{\sum_{i=1}^S 1-a_i} \bigg|  $$
The difference in false-negative rate between the two sensitive attributes,
$$fnr(z_S) =  \bigg| \frac{\sum_{i=1}^S (1 - p_i) y_i a_i}{\sum_{i=1}^S a_i} - \frac{\sum_{i=1}^S(1 -  p_i) y_i (1 - a_i)}{\sum_{i=1}^S 1-a_i}\bigg|$$
Following a similar argument as before the empirical version of EO as defined by Eq. \ref{eq:eo} and also used by \cite{madras18} in the experiments is,
$$const^{EO}(z_S) = fpr + fnr$$ 
EO as defined in \cite{agarwal18} is, 
$$ const^{EO}(z_S) = \max\{fpr , fnr\} $$

Empirical version of DI for a batch of $S$ samples as defined in Eq. \ref{eq:di} as a constraint for binary classes is given by,
$$ const^{DI}(z_S) = - \underset{}{min} \bigg( \frac{ \frac{\sum_{i=1}^S a_i p_i}{\sum_{i=1}^S a_i}} {\frac{\sum_{i=1}^S (1-a_i)
p_i}{\sum_{i=1}^S 1 - a_i}}, \  \frac{\frac{\sum_{i=1}^S (1-a_i) p_i}{\sum_{i=1}^S 1 - a_i} }{ \frac{\sum_{i=1}^S a_i p_i}{\sum_{i=1}^S a_i } } \bigg)$$

The tolerance for each constraint is given by $\epsilon$, which gives the following inequality constraints, for $const^k,\ \  \forall k \in \{DP, EO, DI\} $ the empirical loss for $B$ batches of samples with each batch having $S$ samples denoted by $z_S$,
\begin{equation}
\label{lk:123}
 l_k(h(X), \mathcal{A}, Y): \frac{1}{B}\sum_{l=1}^B const^k(z_S^{(l)}) - \epsilon \leq 0
\end{equation}
Specifically, for  $const^{DI}(z_S)$, $\epsilon = - \frac{p}{100}$, where $p$ is typically set to $80$. 

For maximizing the prediction accuracy, we use cross-entropy loss which is defined as follows for each sample,
$$l_{CE}(h(x_i), y_i) = - y_i \log(p_i) - (1-y_i)\log(1-p_i)$$
The empirical loss, $$\hat{l}_{CE}(h(X), Y) =\frac{1}{SB} \sum_{i=1}^{SB} l_{CE}(h(x_i), y_i)$$
Hence, the overall loss by the Lagrangian method is, 
\begin{equation}
\label{eq:ov1}
L_{NN}(h(X), \mathcal{A}, Y) = \hat{l}_{CE}(h(X), Y) + \lambda \ l_k(h(X), \mathcal{A},Y)
\end{equation}

\subsubsection{Satisfying DP with $Q$-mean loss:} 
The loss due to DP as already defined before is given by Eq. \ref{lk:123}, when $k = DP$.
% \begin{equation}
% \label{lk:dp}
%     l_{DP}(h(X), \mathcal{A}, Y): \frac{1}{B} \sum_{l=1}^{B} const^{DP}(z_S^{(l)}) - \epsilon \leq 0
% \end{equation}
The empirical version of $Q$-mean loss for binary classes that is for $\forall j \in \{0, 1\}$ is defined as follows,
\begin{equation}
\label{eq:qmean}
\sqrt[]{\frac{1}{2}\sum_{i=1}^2 \bigg(1 - \frac{\mathbf{P}(h(x)=i, y=i)}{\mathbf{P}(y = i)} \bigg)^2}
\end{equation}
The corresponding constraint is given by,
\begin{equation*}
    \begin{split}
        &l_Q(h(x_S), y_S) =  \\
        & \sqrt[]{ \bigg( 1 - \frac{\sum_{i=1}^S y_i p_i }{\sum_{i=1}^S y_i}\bigg)^2+ \bigg( 1 - \frac{\sum_{i=1}^S (1-y_i)(1-p_i) }{\sum_{i=1}^S (1-y_i)}\bigg)^2}
    \end{split}{}
\end{equation*}{}
The empirical $Q$-mean loss is, 
$$ \hat{l}_{Q}(h(X), Y) = \frac{1}{B}\sum_{l=1}^B l_Q(h(x_S^{(l)}), y_S^{(l)})$$
Hence, the overall loss by the Lagrangian method is, 
\begin{equation}
\label{eq:ov2}
L_{NN}(h(X), \mathcal{A}, Y) = \hat{l}_{Q}(h(X), Y) + \lambda \ l_{DP}(h(X), \mathcal{A},Y)
\end{equation}
\subsubsection{Lagrangian Multiplier Method}
 The combination of losses and constraints mentioned above are not exhaustive.  The generic definition of the loss could thus be given by, 
 $\forall k \in \{DP, EO, DI\}$
\begin{equation}
\label{eq:overopt}
\boxed{L_{NN} = l_{\theta} + \lambda l_k}
\end{equation} In the equation above, $\lambda$ is the Lagrangian multiplier. Any combination can be tried by changing $l_{\theta}$ and $l_k$ as defined in Eq. \ref{eq:ov1} and Eq. \ref{eq:ov2}. 
The overall optimization of Eq. \ref{eq:overopt} is as follows,
$$\underset{\theta}{min}\  \underset{\lambda}{max} \ L_{NN}$$
The above optimization is carried by performing SGD twice, once for minimizing the loss w.r.t. $\theta$ and again for maximizing the loss w.r.t. $\lambda$ at every iteration \cite{eban16}. 
%%%%%%%%%%%%%%%%%%%%%%%%%%%%%%%%%%%%%%%%%%%%%%%%%%%%%%%%%%%%%
\subsection{Generalization Bounds}
In this subsection, we provide uniform convergence bounds using Rademacher complexity \cite{shalev14} for the loss functions and the constraints discussed above. We assume the class of classifiers learned by the neural network has a finite capacity and we use covering numbers to get this capacity. Given the class of neural network, $\mathcal{H}$, for any $h, \hat{h} \in \mathcal{H}, \ h: \mathbb{R}^d \rightarrow [0,1]$, we define the following $l_{\infty}$ distance: $\max_{x} | h(x) - \hat{h}(x) |$. $\mathcal{N}_{\infty}(\mathcal{H}, \mu)$ is the minimum number of balls of radius $\mu$ required to cover $\mathcal{H}$ under the above distance for any $\mu > 0$. 
\begin{theorem}
\label{thm:gb_1}
For each of $k \in \{DP, EO\}$, the relation between the statistical estimate of the constraint given batches of samples, $z_S$, $\mathbb{E}_{z_S} [const^k(z_S)]$, and the empirical estimate for $B$ batches of samples is listed below. Given that $const^k(z_S) \leq 1$, for a fixed $\delta \in (0,1)$ with a probability at least $1 - \delta$ over a draw of $B$ batches of samples from $(h(X), \mathcal{A}, Y)$, where $h \in \mathcal{H}$,
 
    \begin{equation*}
    \begin{split}
    \mathbb{E}\left[const^{k}(z_s) \right] \leq  \frac{1}{B} \sum_{\ell=1}^{B}  const^{k} & \left(z_S^{(\ell)}\right) 
     + 2 \Omega_{k}  +C  \sqrt{\frac{\log (1 / \delta)}{B}} \\
    \end{split}{}
    \end{equation*}{}
   $$\Omega_{DP, EO} =  \inf _{\mu>0} \biggl\{\mu +  \sqrt{\frac{2 \log \left(\mathcal{N}_{\infty}(\mathcal{H}, \mu/2S)\right)}{B}}\biggr\}$$

Similarly for cross entropy loss $l_{CE}$ and $Q$- mean loss $l_Q$ we get the following bounds.

CE loss: consider  $h(x) = \phi (f(x))$ where $\phi$ is the softmax over the neural network output $f(x)$ where $f\in \mathcal{F}$, assuming $f(x)\leq L$ 
\begin{equation*}
        \begin{split}
 \mathbb{E}[l_{CE}(f(x),y)] \leq \frac{1}{B} \sum_{i=1}^B l_{CE}&(f(x_i),y_i) + 
 2 \Omega_L  +  CL  \sqrt{\frac{\log (1 / \delta)}{B}}\\
        \end{split}{}
    \end{equation*}{}
  $$\Omega_L = \inf _{\mu>0}\biggl\{\mu+  L \sqrt{\frac{2 \log \left(\mathcal{N}_{\infty}(\mathcal{F}, \mu/S)\right)}{B}} \biggr\}$$
 $Q$-mean loss:
    \begin{equation*}
        \begin{split}
\mathbb{E}\left[l_{Q}(h(x_S), y_S) \right] \leq & \frac{1}{B} \sum_{\ell=1}^{B}  l_{Q} \left(h(x_S^{(\ell)}), y_S^{\ell}\right)+ 2 \Omega_Q+ C  \sqrt{\frac{\log (1 / \delta)}{B}} \\
\Omega_Q = \inf _{\mu>0}\biggl\{\mu+ & \sqrt{\frac{2 \log \left(\mathcal{N}_{\infty}(\mathcal{H}, \mu/S)\right)}{B}}\biggr\}
        \end{split}{}
    \end{equation*}{}

 $ C$ is the distribution independent constant.
\end{theorem}{}

\noindent The theorem below gives the bounds for the covering numbers for the class of neural networks that we use for our experiments 

\begin{theorem} \cite{optimal}
\label{thm:gb_2}
For the network with $R$ hidden layers, $D$ parameters, and vector of all model parameters $\parallel w\parallel_1 \leq W $. Given that $w_l$ is bounded, the output of the network is bounded by some constant $L$.
\begin{equation*}
    \begin{split}
        \mathcal{N}_{\infty}(\mathcal{F}, \mu/S) = \mathcal{N}_{\infty}(\mathcal{H}, \mu/S) 
        \leq \left\lceil \frac{DLS(2W)^{R+1}}{\mu} \right\rceil^D
    \end{split}{}
\end{equation*}{}
Hence, on choosing $\mu = \frac{1}{\sqrt{B}}$ we get,
$$  \Omega_{DP} = \Omega_{EO} = \Omega_Q \leq \mathcal{O}\left( \sqrt{RD \frac{\log(WBS DL)}{B}} \right) $$
$$\Omega_{L} = \mathcal{O}\left( L \sqrt{RD \frac{\log(WBS DL)}{B}} \right)$$
\end{theorem}{}
 
\begin{theorem} 
\label{thm:gb_3}
Given $h(x): X \rightarrow [0,1]$, for any $\hat{h}(x): X \rightarrow [0,1]$ such that  $h(x)\neq \hat{h}(x)$, we cannot define a $\widehat{const}_{DI}: (\hat{h}(X), \mathcal{A}, Y) \rightarrow \mathbb{R}$ for a $const_{DI} : (h(X), \mathcal{A}, Y) \rightarrow \mathbb{R}$ such that $|const_{DI}-\widehat{const}_{DI}| \leq \gamma $ is guaranteed, for any $\gamma > 0$. Thus, $\mathcal{N}_{\infty}(\mathcal{DI}, \mu)$ is unbounded for any $\mu>0$ where $\mathcal{DI}$ is set of all possible $const_{DI}$.
\end{theorem}{}

We emphasize that, Theorem \ref{thm:gb_3} indicates that if we approximate DI by a surrogate constraint, however close the learnt classifier is to a desired classifier, the actual DI constraint may get unbounded under specific instances.  That is, even two close classifiers (i.e., $|h(x)-\hat{h}(x) | < \mu $ for any $\mu \in (0,1)$)   may have arbitrarily different DI. For our problem, due to this negative results, we cannot give generalization guarantees by using $\mathcal{N}_{\infty}(\mathcal{DI}, \mu)$ as an upper bound. The few cases where, $DI$ becomes unbounded may not occur in practice as we observe in our experiments that DI results are also comparable. We provide the proofs for all the above theorems in Appendices \ref{appc} \ref{appd}, \ref{appe}.
% as well as there may be possibility of completely another approach to bound $\hat{\mathcal{R}}_L(\mathcal{DI})$.   

While training the network, in the loss we use the $\epsilon$-relaxed fairness constraints as defined in Eq. \ref{lk:123}. We believe that, given the above generalization bounds for the constraints, the trained model will be $\epsilon$-fair with the same bounds.
%%%%%%%%%%%%%%%%%%%%%%%%%%%%%%%%%%%%%%%%%%%%%%%%%%%%%%%%%%%%%%%%%%%%%%%%%%%%%%%%

%%%%%%%%%%%%%%%%%%%%%%%%%%%%%%%%%%%%%%%%%%%%%%%%%%%%%%%%%%%%%%%%%%%%%%%%%%%%%%%%
\section{Experiments and Results}
\label{sec:expt}
In this section, we discuss the network parameters and summarize the results.

\subsection{Hyperparameters}
The architecture that we used is a simple two-layered feed-forward network. The number of hidden neurons in both the layers was one of the following $(100, 50), (200, 100), (500, 100)$. As fairness constraint has no meaning for a single sample, SGD optimizer cannot be used. Hence we use batch sampling. We fix the batch size to be either $1000$ or $500$ depending on the dataset, to get proper estimates of the loss while training. It is to be noted that batch processing is mandatory for this network to be trained efficiently.

% The generic form of softmax function in the output is given by $\frac{e^{(f(x)/t)}}{\sum_{i=1}^m e^{(f(x)/t)}} $, where $t$ is the temperature parameter. The lower the temperature the prediction probabilities are closer to either $1$ or $0$. Since for the complex constraints and losses, we would ideally like the value to be either $0$ or $1$, but hard thresholding would make it non-differentiable. So for our experiments, we set the value of the temperature to enable gradients while simultaneously ensuring a better approximation to the losses or constraints. In our experiments the value of $t$ varies from $ 0.1$ to $0.2$.
For training, we have used the Adam Optimizer with a learning rate of $0.01$ or $0.001$ and the training typically continues for a maximum of $5000$ epochs for each experiment before convergence. The results are averaged over 5-fold cross-validation performance on the data.

% The main challenge in this approach is, it involves many hyper-parameters. It is an exhaustive task to tune all of them. Hyper-parameters are sometimes specific to the dataset used. They also change with the degree of strictness of the constraint enforced. 

\subsection{Performance across datasets}
We have conducted experiments on the six most common datasets used in fairness domain. In Adult, Default, and German dataset, we use gender as the sensitive attribute while predicting income, crime rate, and quality of the customer, respectively, in each of the datasets. In Default and Compass datasets that we used, the race was considered as the sensitive attribute while predicting default payee and recidivism respectively. In the Bank dataset, age is the sensitive attribute while predicting the income of the individual.

\begin{figure}[t!]
    \centering
        \begin{subfigure}[t]{0.22\textwidth}
        \centering
        \includegraphics[width=\textwidth]{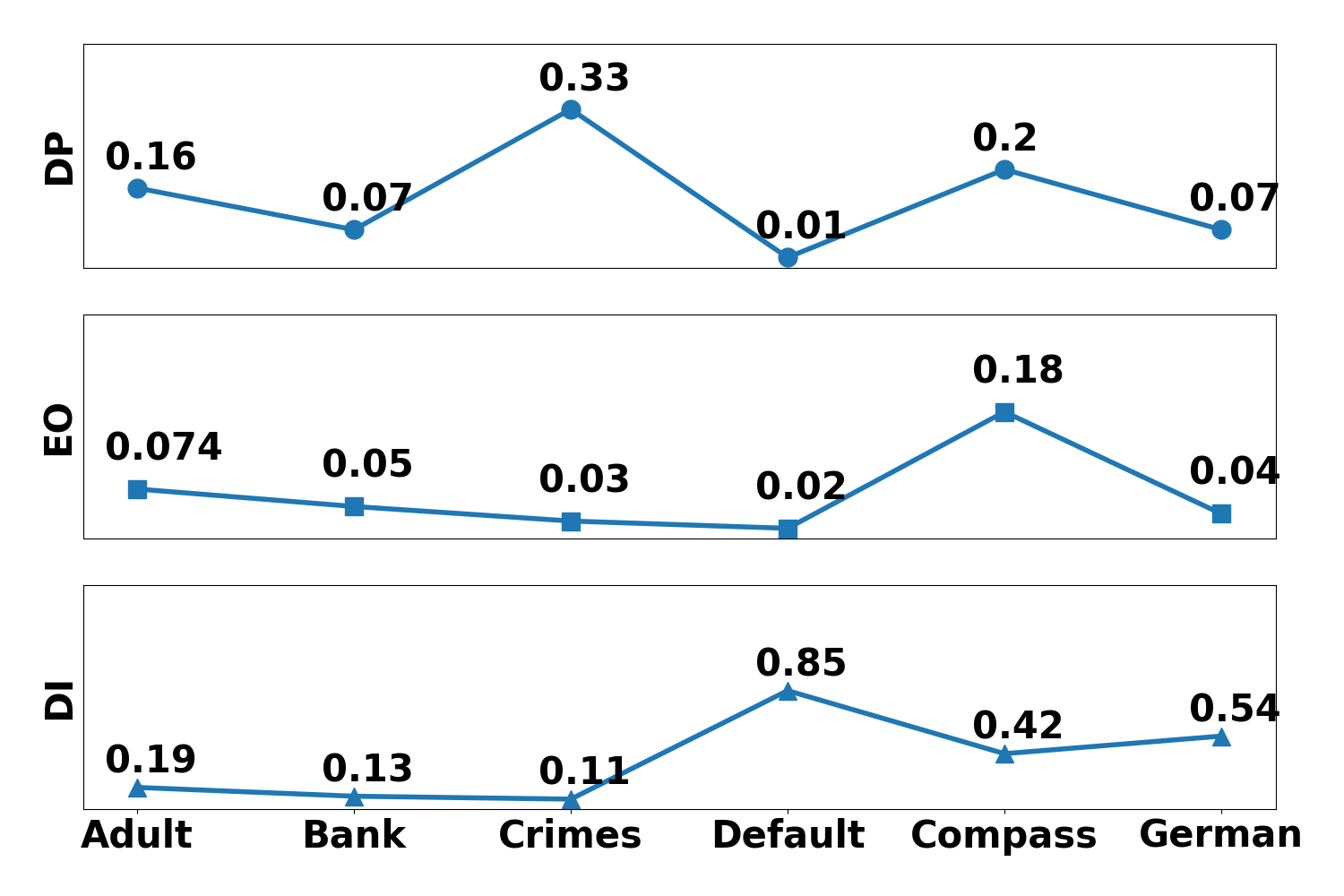}
        \caption{Inherent Bias}
    \end{subfigure}
    ~
    \begin{subfigure}[t]{0.22\textwidth}
        \centering
        \includegraphics[width=\textwidth]{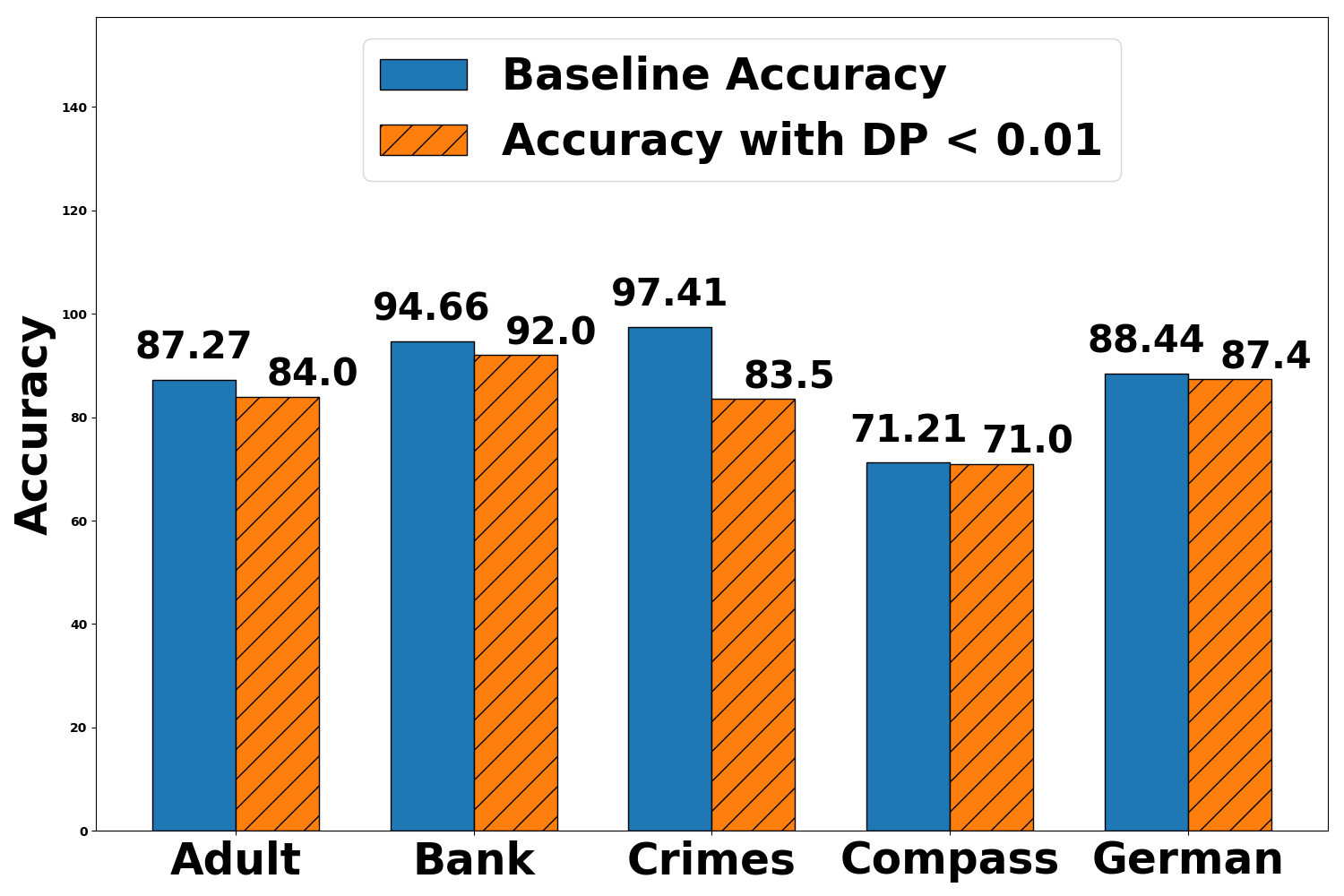}
        \caption{DP}
    \end{subfigure}
    ~ 
    \begin{subfigure}[t]{0.22\textwidth}
        \centering
        \includegraphics[width=\textwidth]{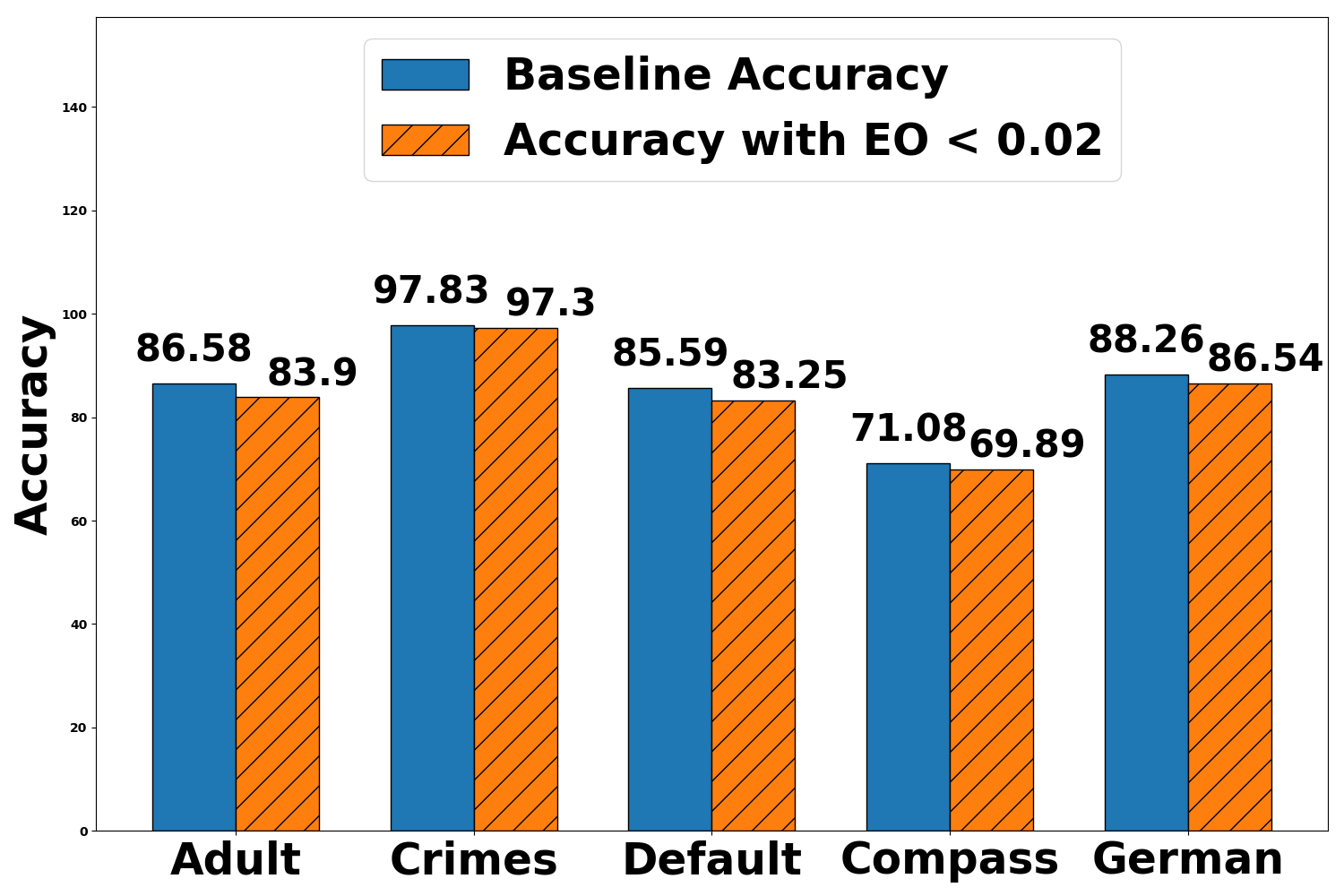}
        \caption{EO}
    \end{subfigure}
    ~
  \begin{subfigure}[t]{0.22\textwidth}
    \centering
    \includegraphics[width=\textwidth]{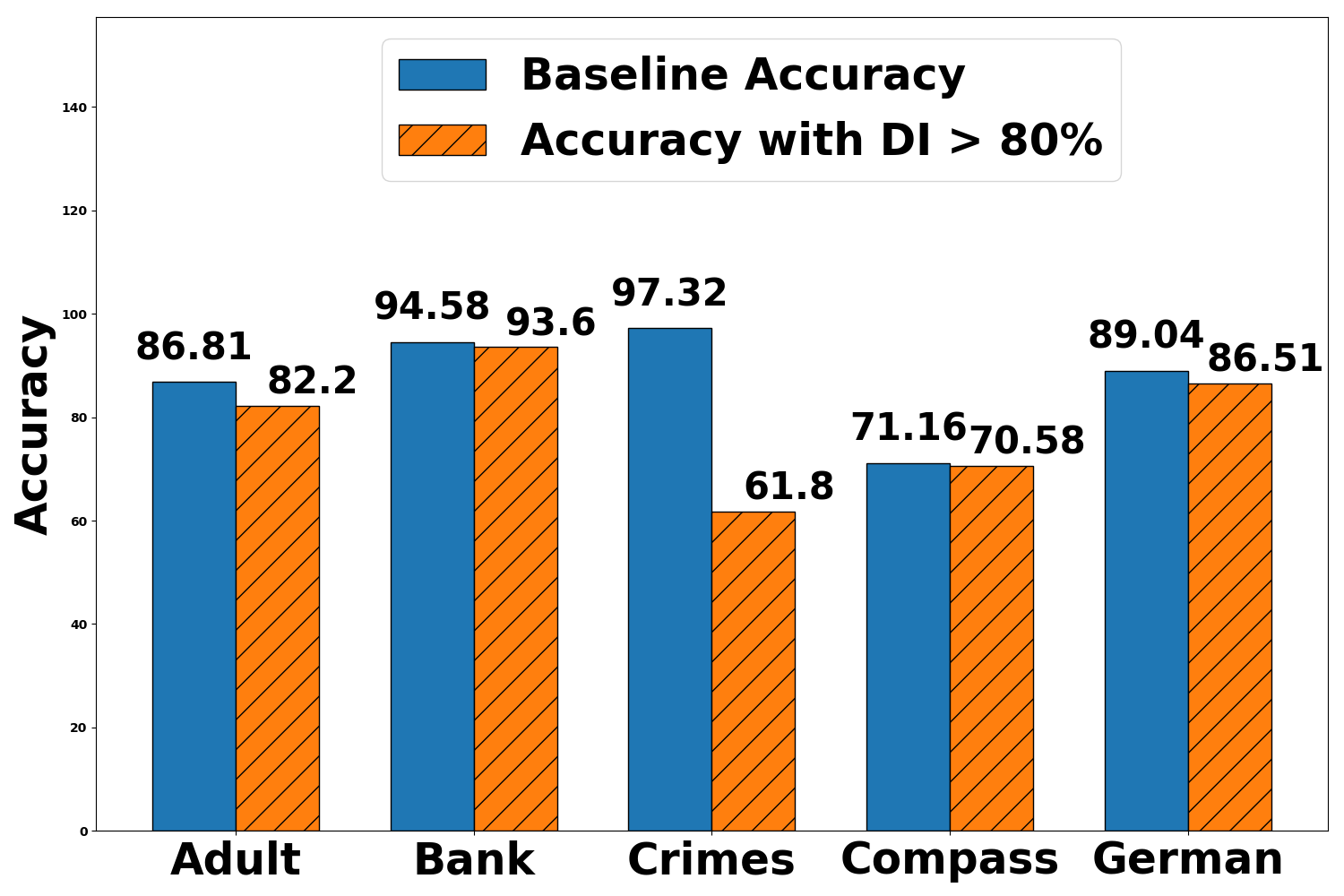}
    \caption{DI}
\end{subfigure}
\caption{Comparison across datasets}
\label{fig:comp}
\end{figure}

In Fig. \ref{fig:comp}a we observe the inherent biases corresponding to each fairness measure within the datasets considered. In order to obtain the values, we set $\lambda=0$ in Eq. \ref{eq:overopt} while training.
% In Fig. \ref{fig:comp}a, the topmost plot we see, that in Crimes, Adult and Compass dataset, DP is violated the most. Violation of EO is highest in Adult and Compass, while DI is violated most in Adult, Bank, and Crimes. 

We compare the baseline accuracy, that is obtained by setting $\lambda=0$, and accuracy using FNNC.
In Fig. \ref{fig:comp}b, we observe a drop in accuracy when the model is trained to limit DP violations within $0.01$, i.e., $\epsilon = 0.01$. There is a significant drop in the accuracy of the Crimes dataset, where the DP is violated the most. Similarly, in Fig. \ref{fig:comp}c and Fig. \ref{fig:comp}d, we study the effect of training the models to limit EO and DI, respectively. We observe that the drop in accuracy is more for datasets that are inherently more biased. In the following section, we compare with other papers and all the results are mostly reported on Adult and Compass dataset. Although all the experiments have single attribute, the approach is easily extendable to multiple attributes.

% \begin{itemize}
% \item Adult: It contains a total of 45,222 samples, each with 14 features. The label is binary which indicates if the income of the sample is above or below 50K USD. Gender is the binary sensitive attribute.
% \item Bank: This dataset has 41,188 data points, each with 20 attributes. The label is binary which indicates whether that person (data point) has subscribed or not to a term deposit. The age is considered as the binary sensitive variable, individuals between 25 and 60 years of age form one group the rest form the other group.
% \item Crimes: There are a total of 2215 individual's data each having 147 features. The label predicts for each individual, whether his crime rate is above average or not. Majority race is the sensitive attribute.
% \item Default: There are a total of 30000 samples with 24 features. Each data sample is a credit user and the task is to predict if the sample would default on payment or not. Gender is considered to be the sensitive attribute.
% \item Compass: The Correctional Offender Management Profiling for Alternative Sanctions, which was made available online by ProPublica. The goal is to predict recidivism within two years. There is data of 5278 individuals each having 9 features. The majority race is used as the protected attribute.
% \item German Dataset: 
% \end{itemize}
\subsection{Comparative Results}

In this subsection, we compare our results with other papers using similar approaches. Instead of reproducing their results, we report the results from their papers directly.

\begin{itemize}[leftmargin=*, noitemsep]
\item\cite{zafar17}: In this paper, the authors propose C-SVM and C-LR to maintain DI while maximizing accuracy. 
We compare our results with theirs on Adult and bank datasets as observed in the Fig. \ref{fig:zafar}. We can see that FNNC obtains higher accuracy for ensuring $p\%$ DI rule for upto $p=80$, for $p > 80$, the accuracy reduces by 2 \%. For obtaining the results we train our network using the loss given in Eq. \ref{lk:123} with $const^{DI}$.
\begin{figure}
 \centering
        \begin{subfigure}[t]{0.4\textwidth}
        \centering
        \includegraphics[width=\textwidth]{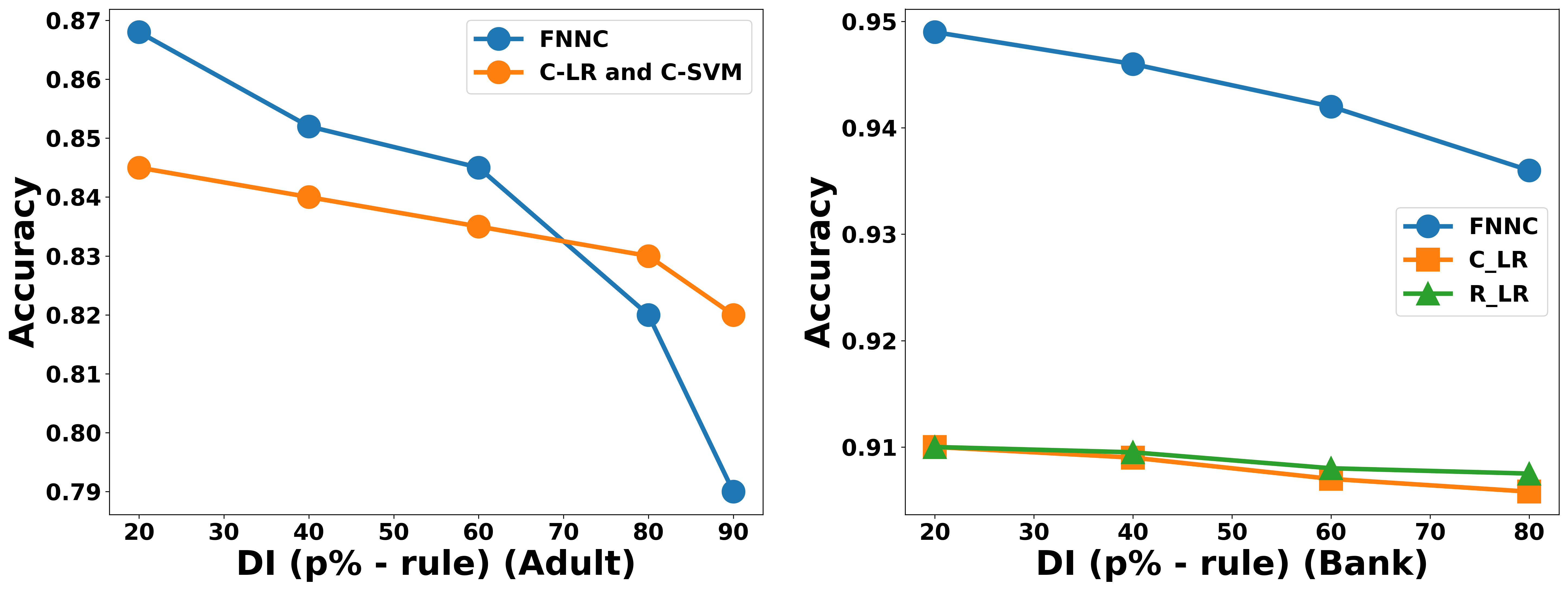}
        \caption{\label{fig:zafar} Accuracy vs $p\%-rule$ comparison of results with Zafar \textit{et al.} on Adult dataset in the left subplot and Bank dataset in the right subplot }
    \end{subfigure}
    ~
    \begin{subfigure}[t]{0.4\textwidth}
        \centering
        \includegraphics[width=\textwidth]{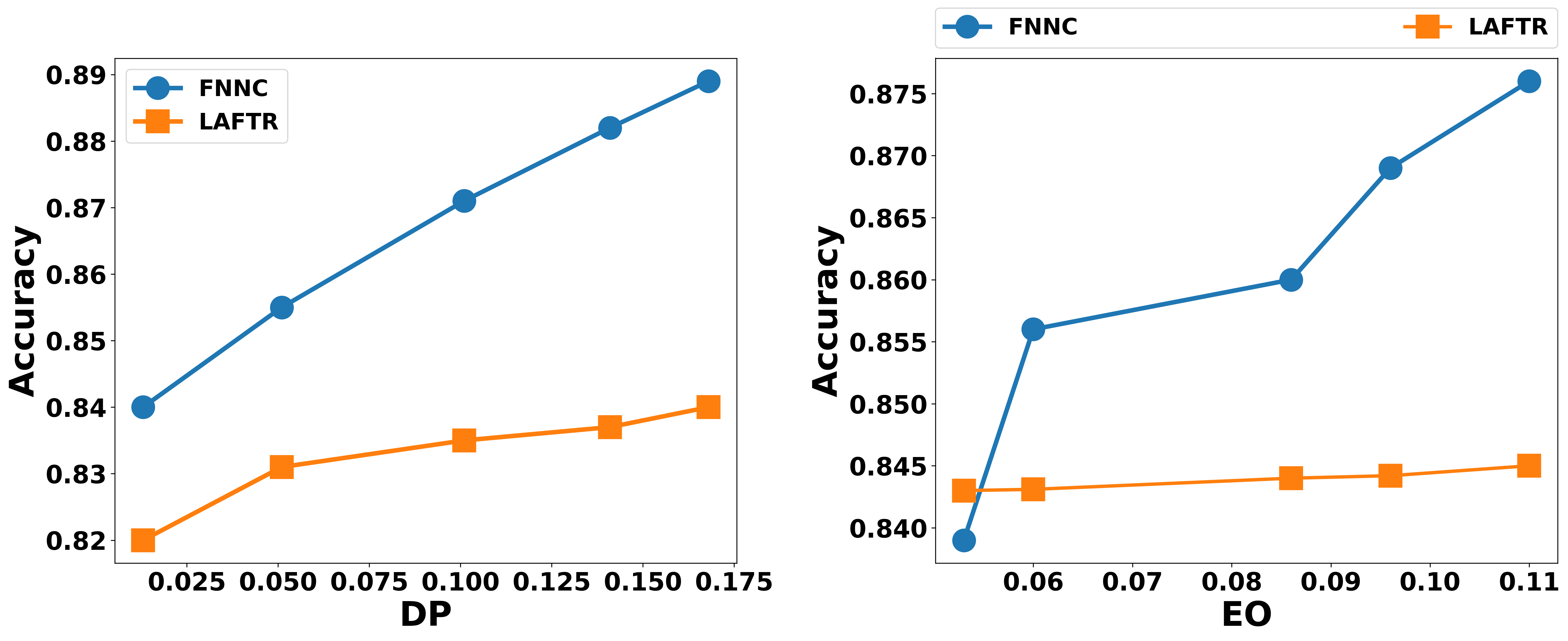}
        \caption{\label{fig:madras}Accuracy vs $\epsilon$ ($\epsilon$ is tolerance for DP and EO respectively) and compare with Madras \textit{et al.} on Adult dataset}
    \end{subfigure}
\caption{Comparative Results}
\end{figure}

\item\cite{madras18}: In this work, the authors propose LAFTR  to ensure DP and EO while maximizing accuracy on Adult dataset. We have compared our results with theirs in Fig. \ref{fig:madras}. For this, we have used loss defined in Eq. \ref{lk:123} with $const^{DP}, const^{EO}$. 

% \begin{figure}[!h]
% \includegraphics[width = 9cm]{adult_eo_dp}
% \caption{\label{fig:madras}Accuracy vs $\epsilon$ ($\epsilon$ is tolerance for DP and EO respectively) and compare with Madras \textit{et al.} on Adult dataset}
% \end{figure}

\begin{figure}[!htb]
 \centering
\includegraphics[width=0.3\textwidth]{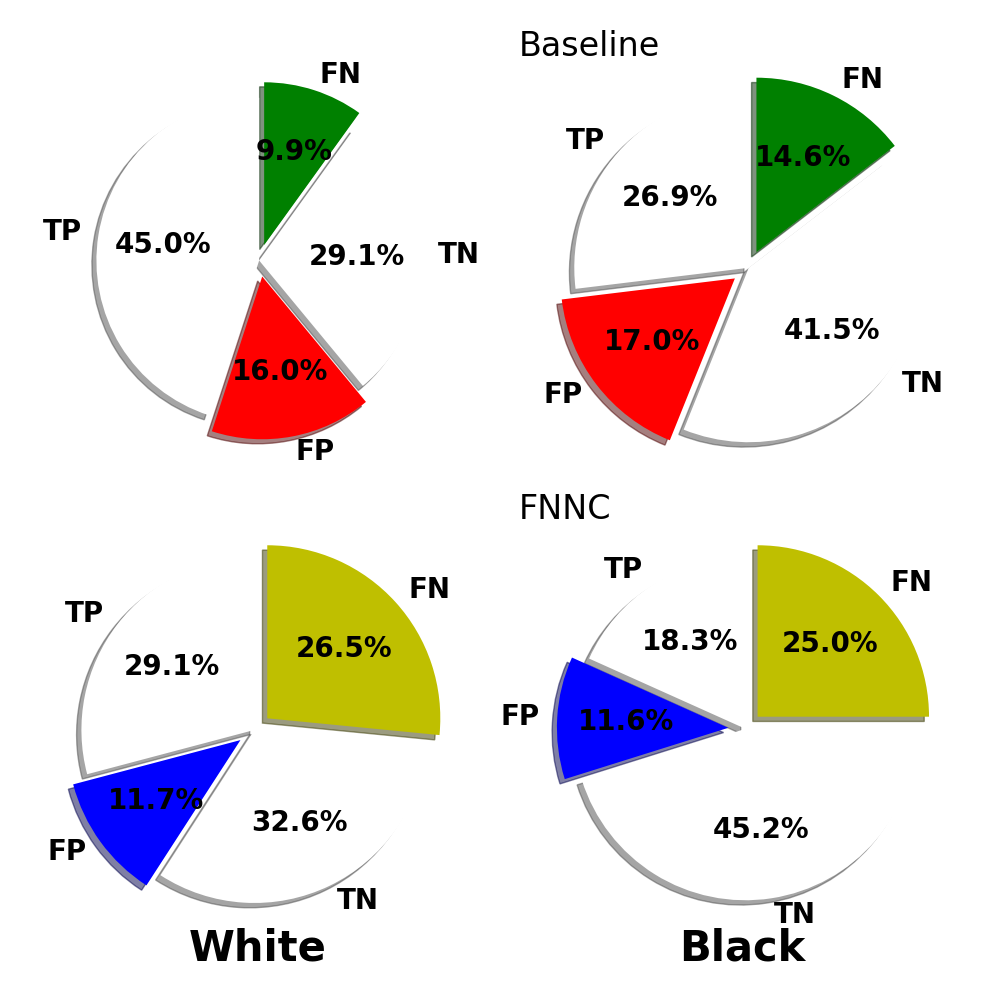}
\caption{\label{fig:pie}Compass dataset:The FPR and FNR is comparable across race in FNNC as observed in the bottom left and right pie charts}
\end{figure}

\begin{table}
\centering
\resizebox{0.5\columnwidth}{!}{%
\renewcommand{\arraystretch}{1.1}
\begin{tabular}{|c|c|c|c|}
\hline
& & Female & Male \\
\hline
% \citeauthor{beutel17} & FPR & 0.0308 & 0.1778 \\
% & FNR & 0.0822 & 0.1520 \\
% \hline
\citeauthor{zhang18} & FPR & 0.0647 & 0.0701 \\
& FNR & 0.4458 & 0.4349 \\
\hline
FNNC & FPR & 0.1228 & 0.1132 \\
& FNR & 0.0797  & 0.0814 \\
\hline
\end{tabular}}
\caption{\label{tab:zhang} False-Positive Rate (FPR) and False-Negative Rate (FNR) for income prediction for the two sex groups in Adult dataset}
\end{table}

\item\cite{zhang18}: The authors  have results for EO on Adult Dataset as can be found in Table \ref{tab:zhang}. Less violation of EO implies that the FPR and FNR values are almost same across different attributes. We get FPR (female) $0.1228$ $\sim$ FPR (male) $0.1132$ and FNR values for female and male are $0.0797$ $\sim$ $0.0814$. The accuracy of the classifier remains at $85\%$. We carry out similar experiments on Compass dataset and compare FNNC with the baseline i.e., trained without fairness constraints in Fig. \ref{fig:pie}

\begin{figure}[!htb]
\centering
\includegraphics[width =8cm]{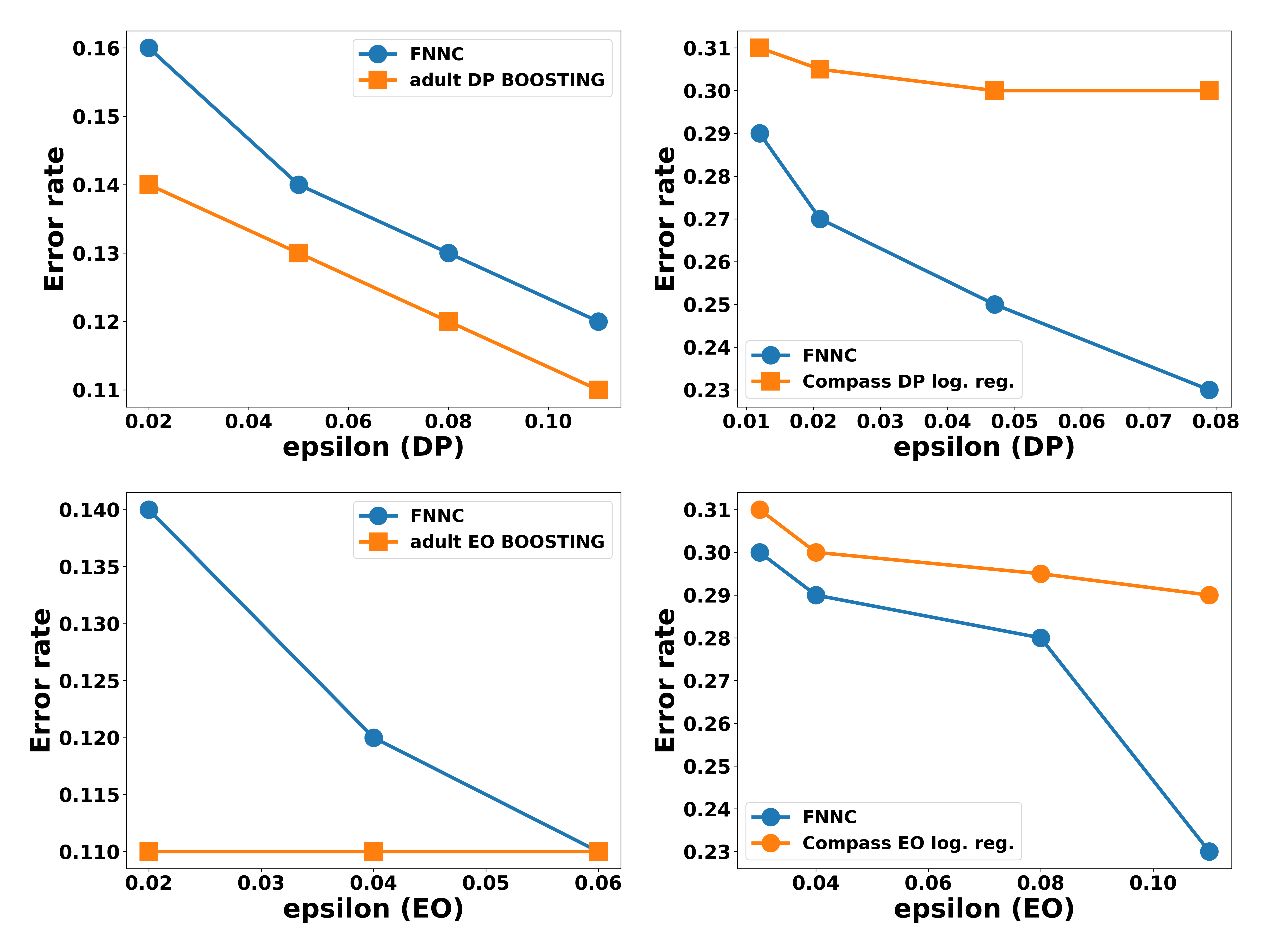}
\caption{\label{fig:agrawal} We compare our results with Agrawal \textit{et al.} for Error rate vs ($\epsilon$) tolerance of DP in top row and EO in bottom row}
\end{figure}

\item\cite{agarwal18}: We compare our results with theirs on Adult and Compass Dataset both for DP and EO as given in Fig. \ref{fig:agrawal}. On observing the plots we find our performance is better for Compass dataset but worse for Adult dataset. The violation of EO in Compass dataset is less compared to the Adult dataset as observed in Fig. \ref{fig:comp}a. Hence, the cost of maintaining fairness is higher in Adult dataset. 
We can observe in Figs. \ref{fig:zafar} \ref{fig:madras}, \ref{fig:agrawal}, that as the fairness constraint is too strict, i.e., $\epsilon$ is very small or $p > 80$, the accuracy reduce or error increases.

\begin{table}
\renewcommand{\arraystretch}{1.1}
\resizebox{\columnwidth}{!}{%
\begin{tabular}{|c|c|c|c|c|}
\hline
Dataset & $\epsilon$ & FNNC & COCO & LinCon \\
\hline
adult   & 0.05 & 0.28 (0.027) & 0.33 (0.035) & 0.39 (0.027) \\
compass & 0.20 & 0.32 (0.147) & 0.41 (0.206) & 0.57 (0.107) \\
crimes  & 0.20 & 0.28 (0.183) & 0.32 (0.197) & 0.52 (0.190) \\
default & 0.05 & 0.29 (0.011) & 0.37 (0.032) & 0.54 (0.015) \\
\hline
\end{tabular}%
}

\caption{\label{tab:qmean} Q-mean loss s.t. DP is within $\epsilon$ (the number in parentheses indicate actual DP)}
\end{table}

\item\cite{narasimhan18}: The authors propose COCO and FRACO algorithm for fractional convex losses with convex constraints. In Table \ref{tab:qmean}, we have results for $Q$-mean loss under DP as the constraint, whose loss function is given by Eq. \ref{eq:ov2}. In the table the values inside the paranthesis correspond to the DP obtained during testing and the values outside the paranthesis is the $Q$-mean loss. We achieve lower $Q$-mean loss than the other approaches on 4 datasets while violation of DP stays within the $\epsilon$.  
\end{itemize}
\section{Discussion and Conclusion}
The results prove that neural networks perform remarkably well on complex and non-convex measures using batch training. From the analysis on generalization bounds, in Theorem \ref{thm:gb_2}, we see that, as $B \rightarrow \infty$, $\Omega \rightarrow 0$. As the number of batches of samples increase, the generalization error asymptotically reduces to zero. The batch size $S$ that we use during the training of the network is a crucial parameter. The generalization error increases in $\sqrt{\log S}$ and also increasing $S$ would reduce $B$ (for fixed data-set). Thus, a smaller value of $S$ is preferable for better generalization. On the other hand, having a very small $S$, would not give a good estimate of the fairness constraint itself. We may end up with sub-optimal classifiers with high loss and less generalization error. Hence, the right balance between $S$ and $B$ is needed to get optimal results.  

We believe that the neural networks can learn optimal feature representation of the data to ensure fairness while maintaining accuracy in an end-to-end manner. Hence, our method, FNNC, combines the traditional approach which learns fair representations by pre-processing the data and the approach for training a fair classifier using surrogate losses. One could consider implementing other non-decomposable performance measures like F1-score, Precision, recall, etc., using this approach, and we leave this for future work.

\bibliographystyle{named}
\bibliography{ref}

\onecolumn
\begin{appendices}
\label{app}
\section{Multi-attribute Case}
The multi-attribute case can be of two different forms as follows,
\begin{itemize}
    \item[a. ]\textit{Multiple binary attributes.} This is a fairly straight forward scenario where we add two different terms of the form of Equation (\ref{lk:123}) corresponding to each attribute.
    \item[b. ] \textit{$m$ groups in single attribute}, $m\geq 2$. We discuss for the DP constraint. For a fixed batch size $S$ of samples given by $z_S=(h(x_S), a_S, y_S)$ and  $p_i = h(x_i) \in [0,1]$. We label the samples of $S$ such that $a^j_i = 1$ if the $i^{th}$ sample belongs to the $j^{th}$ sensitive group  and $a^j_i = 0$ otherwise. \textbf{Note: For binary case, $a_i = 1$ if it belongs to one group (Male) and $0$ otherwise (Female).}
$$const^{DP}_j(z_S) = \bigg|\frac{\sum_{i=1}^S p_i a_i}{\sum_{i=1}^S a_i} - \frac{\sum_{i=1}^S p_i (1 - a_i)}{\sum_{i=1}^S 1-a_i}\bigg| \quad \forall \ j$$
Then we define, 
$$const^{DP}(z_S) = \sum_{j=1}^m const^{DP}_j(z_S)$$
\end{itemize}{}

\section{Note on Convex Surrogates:}
 The fairness constraints are complex and non-convex. One approach to ensure fairness is to introduce convex surrogates to approximate the actual fairness constraints. The convex surrogates are then optimized using existing approaches. Through our experiments and generalization bounds, we claim that FNNC can perform as good as or better than such approaches. The network has the capability of transforming the features into space where the objective is convex, and then stochastic gradient descent helps find the fair and accurate classifier. It is learning the surrogate that leads to the best outcome, which definitely must outperform any fixed surrogate function in some cases, if not all.   
%     \item In the experiments we have compared with the best results reported in the existing work without running them in our own machine. Since we do not do provide an analysis on run time which might be subject to the machine, we considered reporting the values directly from the papers. 
%     \item Although FNNC does not provide much better results comparatively, it provides a simple and easy to implement model as compared to coming up with complex surrogate functions. Moreover, FNNC can be easily modified for newer definitions of fairness as opposed to other approaches. 
%     \item Using deeper network for FNNC led to overfitting. Given the number of samples in the datasets available are fairly small in number and biased.
% \end{itemize}{}

\section{Proof for Theorem \ref{thm:gb_1}}
\label{appc}
\begin{lemma} \cite{rademacher}
\label{lemm:rc}
Let $\mathcal{S} = \{ z_1, \ldots, z_B \}$ be a sample of i.i.d. from some distribution $D$ over $Z$. Then with probability at least $1-\delta$ over a draw of $\mathcal{S}$ from $D$, for all $f \in \mathcal{F}$,
\begin{equation*}
    \begin{split}
        \mathbb{E}_{z\in D}[f(z)] \leq \frac{1}{B} \sum_{i=1}^B f(z_i) + 2 \hat{\mathcal{R}}_B(\mathcal{F}) + 4c\sqrt{\frac{2 \log (4/\delta)}{B}}
    \end{split}{}
\end{equation*}{}
\end{lemma}{}
Given $\hat{\mathcal{R}}_B(\mathcal{F})$ the Rademacher complexity of $\mathcal{F}$, the class of neural network, $\mathcal{H}$, for any $h, \hat{h} \in \mathcal{H}, \ h: \mathbb{R}^d \rightarrow [0,1]$, we define the following $l_{\infty}$ distance: 
\begin{equation}
    \label{eq:dist_h}
    \max_{x} | h(x) - \hat{h}(x) |
\end{equation}{}

\subsubsection{Bounds for DP} 
Given a batch of samples $z_S = (h(x_S), a_S, y_S)$ for a fixed batch size $S$ and $h\in \mathcal{H}$
$$const^{DP}(z_S) = \bigg|\frac{\sum_{i=1}^S h(x_i) a_i}{\sum_{i=1}^S a_i} - \frac{\sum_{i=1}^S h(x_i) (1 - a_i)}{\sum_{i=1}^S 1-a_i}\bigg|$$

Let us consider a class of $\mathcal{DP}$ functions defined on the class of $\mathcal{H}$ as follows,
\begin{equation*}
    \begin{split}
        \mathcal{DP} = \{ const^{DP}: (h(X), \mathcal{A}, Y) & \rightarrow  \mathbb{R} | \ const^{DP}(z_S) = \\ & \bigg|\frac{\sum_{i=1}^S h(x_i) a_i}{\sum_{i=1}^S a_i} - \frac{\sum_{i=1}^S h(x_i) (1 - a_i)}{\sum_{i=1}^S 1-a_i}\bigg| \\ & \text{for some} \ h\in \mathcal{H} \}
    \end{split}{}
\end{equation*}{}

Similar to $\mathcal{N}(\mathcal{H}, \mu)$ given by Eq. \ref{eq:dist_h} we define for $\mathcal{DP}$. Define the $l_{\infty}$ distance as 
$$ \max_{z_S} | \ const^{DP}(z_S) - \widehat{const}^{DP}(z_S) | $$

$\mathcal{N}_{\infty}(\mathcal{DP}, \mu)$ is the minimum number of balls of radius $\mu$ required to cover $\mathcal{DP}$ under the above distance for any $\mu > 0$. 
We apply Lemma \ref{lemm:rc} to the class of demographic parity functions $\mathcal{DP}$. Given a fixed batch size $S$, we have for any $z_S$, $const^{DP}(z_s) \leq 1 $. By definition of the covering number $\mathcal{N}_{\infty}(\mathcal{DP},\mu)$ for any class $const^{DP} \in \mathcal{DP}$, there exists a $\widehat{const}^{DP} \in \hat{\mathcal{DP}}$ where $|\hat{\mathcal{DP}}| \leq \mathcal{N}_{\infty}(\mathcal{DP},\mu)$  such that 
$\max_{z_S} |const^{DP}(z_s) -  \widehat{const}^{DP}| \leq \mu$, for a given $\mu \in (0,1)$
Given $B$ batches of samples where batch is of fixed size $S$

\begin{equation}
\begin{aligned} 
\hat{\mathcal{R}}_{B}(\mathcal{DP}) &=\frac{1}{B} \mathbb{E}_{\sigma}\left[\sup _{const^{DP}} \sum_{\ell=1}^{B} \sigma_{\ell} \cdot  const^{DP}(z_S^{(\ell)}) \right] \\ 
&=\frac{1}{B} \mathbb{E}_{\sigma}\left[\sup _{const^{DP}} \sum_{\ell=1}^{B} \sigma_{\ell} \cdot \widehat{const}^{DP} \left(z_S^{(\ell)}\right)\right]+\frac{1}{B} \mathbb{E}_{\sigma}\left[\sup _{const^{DP}} \sum_{\ell=1}^{B} \sigma_{\ell} \cdot  const^{DP}\left(z_S^{(\ell)}\right)-\widehat{const}^{DP}\left(z_S^{(\ell)}\right)\right]  \\
&\leq \frac{1}{B} \mathbb{E}_{\sigma}\left[ \sup_{\widehat{const}^{DP}} \sum_{\ell=1}^{B} \sigma_{\ell} \cdot \widehat{const}^{DP} \left(z_S^{(\ell)}\right)\right] + \frac{1}{B} \mathbb{E}_{\sigma} \parallel \sigma \parallel_1 \mu \\
& \leq \sqrt{\sum_{\ell}\left( \widehat{const}^{DP}\left(z_S^{\ell}\right)\right)^{2}} \sqrt{\frac{2 \log \left(\mathcal{N}_{\infty}(\mathcal{DP}, \mu)\right)}{B^2}}+\mu \quad \text { (By Massart's Lemma) } \\
& \leq   \sqrt{\frac{2 \log \left(\mathcal{N}_{\infty}(\mathcal{DP}, \mu)\right)}{B}}+\mu
\end{aligned}
\end{equation}
The last inequality is because,

$$ \sqrt{\sum_{\ell}\left( \widehat{const}^{DP}\left(z_S^{\ell}\right)\right)^{2}} \leq \sqrt{\sum_{\ell}\left( const^{DP}\left(z_S^{\ell}\right)+ \mu\right)^{2}} \leq  \sqrt{B} $$

\begin{lemma}
\label{lemm:cover}
$\mathcal{N}_{\infty}(\mathcal{DP}, \mu) \leq \mathcal{N}_{\infty}(\mathcal{H}, \mu/S) $
\end{lemma}{}

\begin{proof}
For any $h, \hat{h} \in \mathcal{H}$ such that for all $x$ we get 
\begin{equation}
    \label{eq:proof1dp}
      | h(x) - \hat{h}(x) | \leq \mu/S
\end{equation}{}

We know that $h(x) \in [0,1] \ \forall h \in \mathcal{H}$. Now let us consider for the class of $\mathcal{DP}$,
\begin{equation*}
    \begin{split}
       |const^{DP} - \widehat{const}^{DP}| &=  \left| \left|\sum_{i=1}^S h(x_i) \left\{ \frac{a_i}{\sum_i a_i} - \frac{1-a_i}{\sum_i 1 - a_i} \right\}\right| - \left|\sum_{i=1}^S \hat{h}(x_i) \left\{ \frac{a_i}{\sum_i a_i} - \frac{1-a_i}{\sum_i 1 - a_i} \right\}\right|  \right| \\
       &\leq \left| \sum_{i=1}^S h(x_i) \left\{ \frac{a_i}{\sum_i a_i} - \frac{1-a_i}{\sum_i 1 - a_i} \right\} - \sum_{i=1}^S \hat{h}(x_i) \left\{ \frac{a_i}{\sum_i a_i} - \frac{1-a_i}{\sum_i 1 - a_i} \right\} \right| \\
       &\leq \left| \sum_{i=1}^S (h(x_i) - \hat{h}({x}_i)) \left\{ \frac{a_i}{\sum_i a_i} - \frac{1-a_i}{\sum_i 1 - a_i} \right\}  \right| \\ 
       &\leq \sum_{i=1}^S \left|(h(x_i) - \hat{h}(x_i)) \left\{ \frac{a_i}{\sum_i a_i} - \frac{1-a_i}{\sum_i 1 - a_i} \right\}  \right|\\
       &\leq \sum_{i=1}^S \left|(h(x_i) - \hat{h}(x_i)) \right| \left|  \frac{a_i}{\sum_i a_i} - \frac{1-a_i}{\sum_i 1 - a_i}  \right| \\
       & \leq \sum_{i=1}^S \left|(h(x_i) - \hat{h}(x_i)) \right| \quad \text{As} \left|  \frac{a_i}{\sum_i a_i} - \frac{1-a_i}{\sum_i 1 - a_i}  \right| \leq 1 \\
       &\leq \mu \quad \text{By Eq. \ref{eq:proof1dp}}
     \end{split}{} 
\end{equation*}
Hence the lemma holds true.
\end{proof}{}

Using the above Lemma \ref{lemm:cover}, we can say that,
$$ \hat{\mathcal{R}}_B(\mathcal{DP}) \leq  \sqrt{\frac{2 \log \left(\mathcal{N}_{\infty}(\mathcal{DP}, \mu)\right)}{B}}+\mu \leq   \sqrt{\frac{2 \log \left(\mathcal{N}_{\infty}(\mathcal{H}, \mu/S)\right)}{B}}+\mu $$

Hence applying the Lemma \ref{lemm:rc}, we get 
$$
\mathbb{E}\left[const^{DP}(z_s) \right] \leq \frac{1}{B} \sum_{\ell=1}^{B}  const^{DP} \left(z_S^{(\ell)}\right)+2 \cdot \inf _{\mu>0}\left\{\mu +  \sqrt{\frac{2 \log \left(\mathcal{N}_{\infty}(\mathcal{H}, \mu/S)\right)}{B}}\right\}+ C  \sqrt{\frac{\log (1 / \delta)}{B}}
$$

\subsubsection{Bounds for EO}
Given a fixed batch size $S$ and $z_S = (h(x_S), a_S, y_S)$,
$$ fpr(z_S) =  \bigg| \frac{\sum_{i=1}^S h(x_i) (1 - y_i) a_i}{\sum_{i=1}^S a_i} - \frac{\sum_{i=1}^S h(x_i)(1 -  y_i)(1 - a_i)}{\sum_{i=1}^S 1-a_i} \bigg|  $$

$$fnr(z_S) =  \bigg| \frac{\sum_{i=1}^S (1 - h(x_i)) y_i a_i}{\sum_{i=1}^S a_i} - \frac{\sum_{i=1}^S(1 -  h(x_i)) y_i (1 - a_i)}{\sum_{i=1}^S 1-a_i}\bigg|$$

$$const^{EO}(z_S) = fpr(z_S) + fnr(z_S)$$ 
As defined in \cite{agarwal18} is, 
$$ const^{EO}(z_S) = \max\{fpr(z_S) , fnr(z_S)\} \leq  fpr(z_S) + fnr(z_S) $$
$$ $$

Let us consider a class of $\mathcal{EO}$ functions defined on the class of $\mathcal{H}$ as follows,
\begin{equation*}
    \begin{split}
        \mathcal{EO} = \{ const^{EO}: (h(X),\mathcal{A}, Y) & \rightarrow  \mathbb{R} | \ const^{EO}(z_S) = \\ &  \bigg| \frac{\sum_{i=1}^S h(x_i) (1 - y_i) a_i}{\sum_{i=1}^S a_i} - \frac{\sum_{i=1}^S h(x_i)(1 -  y_i)(1 - a_i)}{\sum_{i=1}^S 1-a_i} \bigg| \\
        & + \bigg| \frac{\sum_{i=1}^S (1 - h(x_i)) y_i a_i}{\sum_{i=1}^S a_i} - \frac{\sum_{i=1}^S(1 -  h(x_i)) y_i (1 - a_i)}{\sum_{i=1}^S 1-a_i}\bigg| \\ & \text{for some} \ h\in \mathcal{H} \}
    \end{split}{}
\end{equation*}{}

Given $l_{\infty}$ distance as 
$$ \max_{z_S} | \ const^{EO}(z_S) - \widehat{const}^{EO}(z_S) | $$
$\mathcal{N}_{\infty}(\mathcal{EO}, \mu)$ is the minimum number of balls of radius $\mu$ required to cover $\mathcal{EO}$ under the above distance.
We apply Lemma \ref{lemm:rc} to the class of equalized odds class of functions $\mathcal{EO}$. Given a fixed batch size $S$, we have for any $x_S$, $const^{EO}(x_s) \leq 1 $. By definition of the covering number $\mathcal{N}_{\infty}(\mathcal{EO},\mu)$ for any class $const^{EO} \in \mathcal{EO}$, there exists a $\widehat{const}^{EO} \in \hat{\mathcal{EO}}$ where $|\hat{\mathcal{EO}}| \leq \mathcal{N}_{\infty}(\mathcal{EO},\mu)$  such that 
$\max_{x_S} |const^{EO}(x_S) -  \widehat{const}^{EO}| \leq \mu$, for a given $\mu \in (0,1)$.
Given $B$ batches of samples where batch is of fixed size $S$, similar to $\mathcal{DP}$ we can show that, 

\begin{equation}
\hat{\mathcal{R}}_{B}(\mathcal{EO}) \leq   \sqrt{\frac{2 \log \left(\mathcal{N}_{\infty}(\mathcal{EO}, \mu)\right)}{B}}+\mu
\end{equation}

\begin{lemma}
\label{lemm:cover_eo}
$\mathcal{N}_{\infty}(\mathcal{EO}, \mu) \leq \mathcal{N}_{\infty}(\mathcal{H}, \mu/2S) $
\end{lemma}{}

\begin{proof}
For any $h, \hat{h} \in \mathcal{H}$ such that for all $x_S$ we get 
\begin{equation}
    \label{eq:proof1}
      | h(x_i) - \hat{h}(x_i) | \leq \mu/2S
\end{equation}{}

We know that $h(x_i) \in [0,1] \ \forall h \in \mathcal{H}$. Now let us consider for the class of $\mathcal{EO}$,
\begin{equation*}
    \begin{split}
       |const^{EO} - \widehat{const}^{EO}| =& \bigg| \bigg| \frac{\sum_{i=1}^S h(x_i) (1 - y_i) a_i}{\sum_{i=1}^S a_i} - \frac{\sum_{i=1}^S h(x_i)(1 -  y_i)(1 - a_i)}{\sum_{i=1}^S 1-a_i} \bigg| \\
        & + \bigg| \frac{\sum_{i=1}^S (1 - h(x_i)) y_i a_i}{\sum_{i=1}^S a_i} - \frac{\sum_{i=1}^S(1 -  h(x_i)) y_i (1 - a_i)}{\sum_{i=1}^S 1-a_i}\bigg| \\
        & - \bigg| \frac{\sum_{i=1}^S \hat{h}(x_i) (1 - y_i) a_i}{\sum_{i=1}^S a_i} - \frac{\sum_{i=1}^S \hat{h}(x_i)(1 -  y_i)(1 - a_i)}{\sum_{i=1}^S 1-a_i} \bigg| \\
        & - \bigg| \frac{\sum_{i=1}^S (1 - \hat{h}(x_i)) y_i a_i}{\sum_{i=1}^S a_i} - \frac{\sum_{i=1}^S(1 -  \hat{h}(x_i)) y_i (1 - a_i)}{\sum_{i=1}^S 1-a_i}\bigg|\bigg| \\
       \leq & \left| \sum_{i=1}^S h(x_i) \left\{ \frac{a_i (1 - y_i)}{\sum_i a_i} - \frac{(1-a_i)(1 - y_i)}{\sum_i 1 - a_i} \right\} - \sum_{i=1}^S \hat{h}(x_i) \left\{ \frac{a_i (1 - y_i)}{\sum_i a_i} - \frac{(1-a_i)(1 - y_i)}{\sum_i 1 - a_i} \right\} \right| \\
       &+\left| \sum_{i=1}^S (1-h(x_i)) \left\{ \frac{a_i (1 - y_i)}{\sum_i a_i} - \frac{(1-a_i)(1 - y_i)}{\sum_i 1 - a_i} \right\} - \sum_{i=1}^S (1-\hat{h}(x_i)) \left\{ \frac{a_i (1 - y_i)}{\sum_i a_i} - \frac{(1-a_i)(1 - y_i)}{\sum_i 1 - a_i} \right\} \right| \\
       \leq& \left| \sum_{i=1}^S (h(x_i) - h(\hat{x}_i))  \left\{ \frac{a_i (1 - y_i)}{\sum_i a_i} - \frac{(1-a_i)(1 - y_i)}{\sum_i 1 - a_i} \right\}  \right| \\ 
       &+ \left| \sum_{i=1}^S (h(\hat{x}_i) - h(x_i))  \left\{ \frac{a_i (1 - y_i)}{\sum_i a_i} - \frac{(1-a_i)(1 - y_i)}{\sum_i 1 - a_i} \right\}  \right|\\
       \leq & \sum_{i=1}^S 2 \left|(h(x_i) - h(\hat{x}_i)) \left\{ \frac{a_i (1 - y_i)}{\sum_i a_i} - \frac{(1-a_i)(1 - y_i)}{\sum_i 1 - a_i} \right\}  \right|\\
       \leq & \sum_{i=1}^S 2 \left|(h(x_i) - h(\hat{x}_i)) \right| \left|  \frac{a_i (1 - y_i)}{\sum_i a_i} - \frac{(1-a_i)(1 - y_i)}{\sum_i 1 - a_i}   \right| \\
       \leq & \sum_{i=1}^S 2 \left|(h(x_i) - h(\hat{x}_i)) \right| \quad \text{As} \left| \frac{a_i (1 - y_i)}{\sum_i a_i} - \frac{(1-a_i)(1 - y_i)}{\sum_i 1 - a_i}  \right| \leq 1 \\
       \leq &\mu \quad \text{By Eq. \ref{eq:proof1}}
     \end{split}{} 
\end{equation*}
Hence the lemma holds true.
\end{proof}{}

Using the above Lemma \ref{lemm:cover_eo}, we can say that,
$$ \hat{\mathcal{R}}_B(\mathcal{EO}) \leq    \sqrt{\frac{2 \log \left(\mathcal{N}_{\infty}(\mathcal{EO}, \mu)\right)}{B}}+\mu \leq  \sqrt{\frac{2 \log \left(\mathcal{N}_{\infty}(\mathcal{H}, \mu/2S)\right)}{B}}+\mu $$

Hence applying the Lemma \ref{lemm:rc}, we get 
$$
\mathbb{E} \left[const^{EO}(z_S) \right] \leq \frac{1}{B} \sum_{\ell=1}^{B}  const^{EO} \left(z_S^{(\ell)}\right)+2 \cdot \inf _{\mu>0}\left\{\mu+ \sqrt{\frac{2 \log \left(\mathcal{N}_{\infty}(\mathcal{H}, \mu/2S)\right)}{B}}\right\}+ C  \sqrt{\frac{\log (1 / \delta)}{B}}
$$

\subsubsection{Bounds for cross entropy} The loss for a sample $i$ is given by,  $$l_{CE}(h(x_i), y_i) = - y_i \log(h(x_i)) - (1-y_i)\log(1-h(x_i))$$

We have $h(x_i) = \phi (f(x_i))$ where $\phi$ is the softmax over the neural network output $f(x_i)$ where $f\in \mathcal{F}$ 

% \noindent Page 35 Theorem 1.14 \url{https://www-m5.ma.tum.de/foswiki/pub/M5/Allgemeines/MA4801_2016S/ML_notes_main.pdf}
% \url{https://arxiv.org/pdf/1402.3811.pdf}

\begin{lemma}\cite{ledoux}
\label{lemm:radlip}
Let $\mathcal{H}$ be a bounded real-valued function space from some space $\mathcal{Z}$ and $z_1, \ldots, z_n \in \mathcal{Z}$. Let $\xi: \mathbb{R} \rightarrow \mathbb{R}$ be a Lipschitz with constant $L$ and $\xi(0) = 0$. Then, we have
$$E_{\sigma} \sup _{h \in \mathcal{H}} \frac{1}{n} \sum_{i \in[n]} \sigma_{i} \xi\left(h\left(\mathbf{z}_{i}\right)\right) \leq L E_{\sigma} \sup _{h \in \mathcal{H}} \frac{1}{n} \sum_{i \in[n]} \sigma_{i} h\left(\mathbf{z}_{i}\right)$$
\end{lemma}{}

\begin{lemma}
\label{lemm:radlip_1}
$l_{CE}(.,.)$ is $L$ Lipschitz with first argument  hence, 
    $$ \hat{\mathcal{R}}_{L}(\mathcal{CE} \circ \mathcal{F}) \leq   \hat{\mathcal{R}}_{L}(\mathcal{F})  $$
where $\mathcal{CE} = \{l_{CE}(f(x),y)| \ \forall f \in \mathcal{F}\}$
\end{lemma}{}
\begin{proof}
Given that 
$$l_{CE}(f(x), y) =  y_i \log(y/ \phi(f(x_i)))+  (1-y_i) \log((1- y_i)/ (1 - \phi(f(x_i)))) $$

It is easy to find that $\partial l_{CE}(f(x), y) / \partial f(x) \in[-1,1]^m$ and thus $l_{CE}$ is a $1$-Lipschitz function with its first argument. Given $l^{\prime}_{CE}(.,.) = l_{CE}(.,.) - l_{CE}(0,.)$ and we can get that $\hat{\mathcal{R}}(l_{CE}\circ f) = \hat{\mathcal{R}}(l^{\prime}_{CE}\circ f)$, then we apply Lemma \ref{lemm:radlip} to $l^{\prime}_{CE}$ and conclude the proof.
\end{proof}{}
From Lemma \ref{lemm:rc} and \ref{lemm:radlip_1}, we obtain the following,
\begin{equation}
\label{eq:ce_proof}
    \mathbb{E}_{x\in \mathcal{X}}[l_{CE}(f(x),y)] \leq \frac{1}{B} \sum_{i=1}^B l_{CE}(f(x_i),y_i) + 2 \hat{\mathcal{R}}_B(\mathcal{F}) + 4c\sqrt{\frac{2 \log (4/\delta)}{B}}
\end{equation}{}

From the above Eq. \ref{eq:ce_proof} we need to compute $\hat{\mathcal{R}_L}(\mathcal{F})$. Given any sample $x$, $f(x) \leq L $. By definition of the covering number $\mathcal{N}_{\infty}(\mathcal{F},\mu)$ for any class $f \in \mathcal{F}$, there exists a $\widehat{f} \in \hat{\mathcal{F}}$ where $|\hat{\mathcal{F}}| \leq \mathcal{N}_{\infty}(\mathcal{F},\mu)$  such that 
$\max_{x} |f(x) -  \widehat{f}| \leq \mu$, for a given $\mu \in (0,1)$.
Given $B$ samples,

\begin{equation}
\begin{aligned} 
\hat{\mathcal{R}}_{B}(\mathcal{F}) &=\frac{1}{B} \mathbb{E}_{\sigma}\left[\sup _{f\in \mathcal{F}} \sum_{\ell=1}^{B} \sigma_{\ell} \cdot f(x^{(\ell)}) \right] \\ 
&=\frac{1}{B} \mathbb{E}_{\sigma}\left[\sup _{f\in \mathcal{F}} \sum_{\ell=1}^{B} \sigma_{\ell} \cdot \widehat{f} \left(x^{(\ell)}\right)\right]+\frac{1}{B} \mathbb{E}_{\sigma}\left[\sup _{f\in \mathcal{F}} \sum_{\ell=1}^{B} \sigma_{\ell} \cdot  f\left(x_S^{(\ell)}\right)-\widehat{f}\left(x^{(\ell)}\right)\right]  \\
&\leq \frac{1}{B} \mathbb{E}_{\sigma}\left[ \sup_{\widehat{f}} \sum_{\ell=1}^{B} \sigma_{\ell} \cdot \widehat{f} \left(x^{(\ell)}\right)\right] + \frac{1}{B} \mathbb{E}_{\sigma} \parallel \sigma \parallel_1 \mu \\
& \leq \sqrt{\sum_{\ell}\left( \widehat{f}\left(x_S^{\ell}\right)\right)^{2}} \sqrt{\frac{2 \log \left(\mathcal{N}_{\infty}(\mathcal{F}, \mu)\right)}{B^2}}+\mu \quad \text { (By Massart's Lemma) } \\
& \leq   L \sqrt{\frac{2 \log \left(\mathcal{N}_{\infty}(\mathcal{F}, \mu)\right)}{B}}+\mu
\end{aligned}
\end{equation}
The last inequality is because,
$$ \sqrt{\sum_{\ell}\left( \widehat{f}\left(x_S^{\ell}\right)\right)^{2}} \leq \sqrt{\sum_{\ell}\left( f\left(x_S^{\ell}\right)+ \mu\right)^{2}} \leq  L\sqrt{B} $$
Hence,
$$ \hat{\mathcal{R}}_B(\mathcal{F}) \leq  L\sqrt{\frac{2 \log \left(\mathcal{N}_{\infty}(\mathcal{F}, \mu)\right)}{B}}+\mu $$

Hence applying the Lemma \ref{lemm:rc}, we get 
$$
 \mathbb{E}[l_{CE}(f(x),y)] \leq \frac{1}{B} \sum_{i=1}^B l_{CE}(f(x_i),y_i) + 2 \cdot \inf _{\mu>0}\left\{\mu+  L \sqrt{\frac{2 \log \left(\mathcal{N}_{\infty}(\mathcal{F}, \mu/S)\right)}{B}}\right\}+ CL  \sqrt{\frac{\log (1 / \delta)}{B}}
$$

\subsubsection{Bounds for $Q$-mean loss:} 
Given batch of size $S$, having samples $(x_S, y_S)$. 

Here we prove for when the dataset has $m$ different classes, but for us $m=2$ 
$$   l_Q(h(x_S), y_S) = \sqrt[]{\frac{1}{m} \sum_{j=1}^m \bigg( 1 - \frac{\sum_{i=1}^S y_i^j h^j(x_i) }{\sum_{i=1}^S y_i^j}\bigg)^2} $$

Let us consider a class of $\mathcal{Q}$ functions defined on the class of $\mathcal{H}$ as follows,
\begin{equation*}
    \begin{split}
        \mathcal{Q} = \{ l_{Q}: (X, Y) & \rightarrow  \mathbb{R} | \ l_{Q}(x_S) = \\ &  \sqrt[]{\frac{1}{m} \sum_{j=1}^m \bigg( 1 - \frac{\sum_{i=1}^S y_i^j h^j(x_i) }{\sum_{i=1}^S y_i^j}\bigg)^2} \\ & \text{for some} \ h\in \mathcal{H} \}
    \end{split}{}
\end{equation*}{}

Define the $l_{\infty, 1}$ distance as 
$$ \max_{x_S} \parallel \ l_{Q}(x_S) - \widehat{l}_{Q}(x_S) \parallel $$
$\mathcal{N}_{\infty}(\mathcal{Q}, \mu)$ is the minimum number of balls of radius $\mu$ required to cover $\mathcal{Q}$ under the above distance.

Let us now apply the Lemma \ref{lemm:rc} to the class of functions $\mathcal{Q}$. Given a fixed batch size $S$, we have for any $x_S, y_S$, $l_{Q}(x_S, y_S) \leq 1 $. By definition of the covering number $\mathcal{N}_{\infty}(\mathcal{Q},\mu)$ for any class $l_{Q} \in \mathcal{Q}$, there exists a $\widehat{l}_{Q} \in \hat{\mathcal{Q}}$ where $|\hat{\mathcal{Q}}| \leq \mathcal{N}_{\infty}(\mathcal{Q},\mu)$  such that 
$\max_{(x_S, y_S)} |l_{Q}(x_S, y_S) -  \widehat{l}_{Q}(x_S, y_S)| \leq \mu$, for a given $\mu \in (0,1)$.
Given $B$ batches of samples where batch is of fixed size $S$, similar to $\mathcal{DP}$ we can show that, 

\begin{equation}
\hat{\mathcal{R}}_{B}(\mathcal{Q}) \leq   \sqrt{\frac{2 \log \left(\mathcal{N}_{\infty}(\mathcal{Q}, \mu)\right)}{B}}+\mu
\end{equation}

\begin{lemma}
\label{lemm:cover_qmean}
$\mathcal{N}_{\infty}(\mathcal{Q}, \mu) \leq \mathcal{N}_{\infty}(\mathcal{H}_S, \mu/S) $
\end{lemma}{}

\begin{proof}
We know that $h(x_i) \in [0,1]^m \ \forall h \in \mathcal{H}$. For any $h, \hat{h} \in \mathcal{H}$ such that for all $x_S$ we get, the following $l_{\infty, 1}$ 
\begin{equation}
    \label{eq:proof2dp}
      \parallel h(x_i) - \hat{h}(x_i) \parallel \leq \mu/S
\end{equation}{}

 Now let us consider for the class of $\mathcal{Q}$, 
\begin{equation*}
    \begin{split}
       |const^{Q} - \widehat{const}^{Q}| &=  \left|  \sqrt[]{\frac{1}{m} \sum_{j=1}^m \bigg( 1 - \frac{\sum_{i=1}^S y_i^j h^j(x_i) }{\sum_{i=1}^S y_i^j}\bigg)^2} -  \sqrt[]{\frac{1}{m} \sum_{j=1}^m \bigg( 1 - \frac{\sum_{i=1}^S y_i^j \hat{h}^j(x_i) }{\sum_{i=1}^S y_i^j}\bigg)^2}  \right| \\
      & \leq \frac{1}{\sqrt{m}}  \left|  \sqrt[]{ \sum_{j=1}^m \bigg( 1 - \frac{\sum_{i=1}^S y_i^j h^j(x_i) }{\sum_{i=1}^S y_i^j}\bigg)^2} -  \sqrt[]{\sum_{j=1}^m \bigg( 1 - \frac{\sum_{i=1}^S y_i^j \hat{h}^j(x_i) }{\sum_{i=1}^S y_i^j}\bigg)^2}  \right| \\
       & \leq \frac{1}{\sqrt{m}}   \sqrt[]{ \sum_{j=1}^m  \left| \frac{\sum_{i=1}^S y_i^j \hat{h}^j(x_i) }{\sum_{i=1}^S y_i^j} - \frac{\sum_{i=1}^S y_i^j h^j(x_i) }{\sum_{i=1}^S y_i^j} \right|^2}   \quad \text{Triangle Inequality}\\
       & \leq \frac{1}{\sqrt{m}}   \sum_{j=1}^m  \left| \frac{\sum_{i=1}^S y_i^j \hat{h}^j(x_i) }{\sum_{i=1}^S y_i^j} - \frac{\sum_{i=1}^S y_i^j h^j(x_i) }{\sum_{i=1}^S y_i^j} \right|   \\
       & \leq \frac{1}{\sqrt{m}} \sum_{j=1}^m\left| \sum_{i=1}^S y_i^j (\hat{h}^j(x_i)-  h^j(x_i) ) \right| \quad \text{As  $\sum_{i=1}^S y_i^j \geq 1 $}\\
       &\leq  \sum_{i=1}^S \sum_{j=1}^m  \left|  y_i^j (\hat{h}^j(x_i)-  h^j(x_i) ) \right| \\
       & \leq \sum_{i=1}^S \sum_{j=1}^m \left| y_i^j (\hat{h}^j(x_i)-  h^j(x_i) ) \right| \\
       & \leq \sum_{i=1}^S \parallel \hat{h}^j(x_i)-  h^j(x_i) \parallel \quad \text{As $y_i^j \leq 1$} \\
       &\leq \mu \quad \text{By Eq. \ref{eq:proof2dp}}
     \end{split}{} 
\end{equation*}
Hence the lemma holds true.
\end{proof}{}

Using the above Lemma \ref{lemm:cover_qmean}, we can say that,
$$ \hat{\mathcal{R}}_B(\mathcal{Q}) \leq  \sqrt{\frac{2 \log \left(\mathcal{N}_{\infty}(\mathcal{Q}, \mu)\right)}{B}}+\mu \leq  \sqrt{\frac{2 \log \left(\mathcal{N}_{\infty}(\mathcal{H}, \mu/S)\right)}{B}}+\mu $$

Hence applying the Lemma \ref{lemm:rc}, we get 
$$
\mathbb{E}\left[l_{Q}(x_S,y_S) \right] \leq \frac{1}{B} \sum_{\ell=1}^{B}  l_{Q} \left(x_S^{(\ell)}, y_S^{(\ell)}\right)+2 \cdot \inf _{\mu>0}\left\{\mu+ \sqrt{\frac{2 \log \left(\mathcal{N}_{\infty}(\mathcal{H}, \mu/S)\right)}{B}}\right\}+ C  \sqrt{\frac{\log (1 / \delta)}{B}}
$$

\section{Proof for Theorem \ref{thm:gb_2}}
\label{appd}
\begin{lemma} \cite{optimal}
\label{lemm:nn} 
Let $\mathcal{H}_k$ be a class of feed-forward neural networks that maps an input vector $x \in \mathbb{R}^{d} $ to an output vector $o \in \mathbb{R}$, with each layer $l$ containing $T_l$ nodes and computing $z \mapsto \phi_{l}(w^l z)$ where each $w^l \in \mathbb{R}^{T_l \times T_{l-1}} $ and $\phi_l: \mathbb{R}^{T_l} \rightarrow [-L, +L]^{T_l}$. Further let, for each network in $\mathcal{F}_k$, let the parameters $\parallel w^l \parallel_1 \leq W$ and $\parallel \phi_l(s) - \phi_l(s^{\prime}) \parallel \leq \Phi \parallel s - s^{\prime} \parallel $ for any $s, s^{\prime} \in \mathbb{R}^{T_{l-1}}$
$$ \mathcal{N}_{\infty}(\mathcal{F}_k, \mu) \leq \left\lceil\frac{2LD^2W(2\Phi W)^k}{\mu}\right\rceil^D $$
where $D$ is the total number of parameters
\end{lemma}{}

\noindent Using the above lemma we prove the Theorem \ref{thm:gb_2},
\begin{proof}
The architecture that we use are 2 layered feed-forward neural networks with at most $K$ hidden nodes per layer. For each layer $l$ we assume, the $\parallel w_l \parallel_1 \leq W $. We know that ReLU activation and softmax activation are $1$-Lipschitz \cite{optimal}. Given that the input $X$ has $d$ dimensions and $w_l$ is bounded, the output of ReLU is bounded by some constant $L$.
By applying Lemma \ref{lemm:nn} with $\Phi = 1$, 
$$ \mathcal{N}_{\infty}(\mathcal{H}, \mu/S) \leq \left\lceil \frac{DLS(2W)^{R+1}}{\mu} \right\rceil^D$$
Hence, on choosing $\mu = \frac{1}{\sqrt{B}}$ we get,

\begin{equation*}
    \begin{split}
         \Omega & \leq  \frac{1}{\sqrt{B}} + \sqrt{\frac{2 \log(\lceil(DLS(2W)^{R+1} B^{1/2} \rceil^D  )}{B} } \\
        & \leq   \mathcal{O}\left( \sqrt{ \frac{RD \log(WBSDL)}{B}} \right)
    \end{split}{}
\end{equation*}{}
where $\Omega = \{\Omega_{DP}, \Omega_{EO}, \Omega_{Q}\} $, similarly proof works for $\Omega_{L}$
\end{proof}{}

\section{Proof for Theorem \ref{thm:gb_3}}
\label{appe}
\begin{lemma}
\label{lemm:di} Given $a, b \geq 0$, 
$|\min(a, \frac{1}{a}) - \min(b, \frac{1}{b})| \leq |a - b|$
\end{lemma}{}
\begin{proof}
It trivially holds true when,  
\begin{itemize}
    \item CASE 1: $\min(a, \frac{1}{a}) = a, \ \min(b, \frac{1}{b}) = b$
    \item CASE 2: $\min(a, \frac{1}{a}) = \frac{1}{a}, \ \min(b, \frac{1}{b}) = b$
\end{itemize}{}
Let us consider the following cases,
\begin{itemize}
    \item CASE 3: $\min(a, \frac{1}{a}) = a, \ \min(b, \frac{1}{b}) = \frac{1}{b}$ \\
    We know that $a \leq 1$ hence, $2a \leq b + \frac{1}{b}$ which gives that $a - \frac{1}{b} \leq b - a$. for this case $a - b \leq a - \frac{1}{b}$, hence $|a - \frac{1}{b}| \leq |a - b|$
    \item CASE 4: $\min(a, \frac{1}{a}) = \frac{1}{a}, \ \min(b, \frac{1}{b}) = \frac{1}{b}$ \\
    In this case $|\frac{1}{a} - \frac{1}{b}| \leq |\frac{b - a}{ab}| \leq |a - b|$  as $a, b \geq 1$
\end{itemize}{}
\end{proof}{}

\noindent Using the above lemma we prove the Theorem \ref{thm:gb_3},
\begin{proof}
For any $h, \hat{h} \in \mathcal{H}$ such that for all $x_S$ we get 
\begin{equation}
    \label{eq:proof_di}
      | h(x_i) - \hat{h}(x_i) | \leq \mu
\end{equation}{}

 We assume that $S = 100$, $\sum_{i=1}^S a_i = 50$ and $\sum_{i=1}^S 1 - a_i = 50$, for $a_i = 1, \hat{h}(x)=1$ and $h(x) = 1$. For $a_i = 0, \hat{h}(x) = \mu$ and $h(x) = \delta$ where $\delta \in (0,1)$ s.t $|\mu - \delta| \leq \mu$. Now let us consider for the class of $\mathcal{DI}$,
\begin{equation*}
    \begin{split}
       |const^{DI} - \widehat{const}^{DI}| =&\bigg|  \underset{}{min} \bigg( \frac{ \frac{\sum_{i=1}^S a_i h(x_i)}{\sum_{i=1}^S a_i}} {\frac{\sum_{i=1}^S (1-a_i)
h(x_i)}{\sum_{i=1}^S 1 - a_i}}, \  \frac{\frac{\sum_{i=1}^S (1-a_i) h(x_i)}{\sum_{i=1}^S 1 - a_i} }{ \frac{\sum_{i=1}^S a_i h(x_i)}{\sum_{i=1}^S a_i } } \bigg) -  \underset{}{min} \bigg( \frac{ \frac{\sum_{i=1}^S a_i \hat{h}(x_i)}{\sum_{i=1}^S a_i}} {\frac{\sum_{i=1}^S (1-a_i)
\hat{h}(x_i)}{\sum_{i=1}^S 1 - a_i}}, \  \frac{\frac{\sum_{i=1}^S (1-a_i) \hat{h}(x_i)}{\sum_{i=1}^S 1 - a_i} }{ \frac{\sum_{i=1}^S a_i \hat{h}(x_i)}{\sum_{i=1}^S a_i } } \bigg) \bigg| \\
\leq & \bigg|  \frac{ \frac{\sum_{i=1}^S a_i h(x_i)}{\sum_{i=1}^S a_i}} {\frac{\sum_{i=1}^S (1-a_i)
h(x_i)}{\sum_{i=1}^S 1 - a_i}} - \frac{ \frac{\sum_{i=1}^S a_i \hat{h}(x_i)}{\sum_{i=1}^S a_i}} {\frac{\sum_{i=1}^S (1-a_i)
\hat{h}(x_i)}{\sum_{i=1}^S 1 - a_i}} \bigg| \quad \text{By Lemma \ref{lemm:di}}\\
\leq & \bigg|\frac{\sum_{i=1}^S (1-a_i)}{\sum_{i=1}^S a_i}\frac{\sum_{i=1}^S a_i h(x_i)}{\sum_{i=1}^S (1-a_i)h(x_i)} - \frac{\sum_{i=1}^S (1-a_i)}{\sum_{i=1}^S a_i}\frac{\sum_{i=1}^S a_i \hat{h}(x_i)}{\sum_{i=1}^S (1-a_i)\hat{h}(x_i)} \bigg| \\
\leq & \bigg| \frac{\sum_{i=1}^S (1-a_i)}{\sum_{i=1}^S a_i} \bigg( \frac{\sum_{i=1}^S a_i h(x_i)}{\sum_{i=1}^S (1-a_i)h(x_i)} - \frac{\sum_{i=1}^S a_i \hat{h}(x_i)}{\sum_{i=1}^S (1-a_i)\hat{h}(x_i)}  \bigg) \bigg| \\
\leq &  \left| \frac{50}{50 \delta} - \frac{50 }{50 \mu }\right|\\
\leq &  \left| \frac{1}{ \delta} - \frac{1}{ \mu } \right|\\
 \end{split}{} 
\end{equation*}
Given a fixed $\mu$, we can have an arbitrarily small $\delta$ such that the above becomes unbounded, hence the theorem follows.
\end{proof}{}

\end{appendices}

\end{document}